\documentclass[11pt]{article}

\usepackage[preprint]{acl}
\usepackage{times}
\usepackage{latexsym}
\usepackage[T1]{fontenc}
\usepackage[utf8]{inputenc}
\usepackage{microtype}
\usepackage{inconsolata}
\usepackage{graphicx}
\usepackage{booktabs}
\usepackage{amsmath}
\usepackage{placeins}
\usepackage{flafter}
\usepackage{authblk}

\def\secref#1{\S\ref{sec:#1}}
\def\seclabel#1{\label{sec:#1}}

\newcommand{\pixel}{PIXEL}

\title{Beyond Clean Accuracy: The Practical Challenges and Benefits of Pixel-Based Language Models}

\author[]{\textbf{Haowen Tang}$^{1}$}
\author[]{\textbf{Wei Xia}$^{1}$}
\author[]{\textbf{Raoyuan Zhao}$^{1,2}$}
\author[]{\textbf{Hinrich Sch\"utze}$^{1,2}$}
\author[]{\textbf{Yihong Liu}$^{1,2,\dagger}$}

\affil[]{$^1$Center for Information and Language Processing, LMU Munich \\$^2$Munich Center for Machine Learning (MCML)
 \protect\\
  \texttt{\{Haowen.Tang, W.Xia1\}@campus.lmu.de}
 \space \space \space
 \texttt{\{rzhao, yihong\}@cis.lmu.de}} 

\begin{document}
\maketitle

\def\thefootnote{$\dag$}\footnotetext{Corresponding author.}\def\thefootnote{\arabic{footnote}}

\begin{abstract}
Pixel-based language models process rendered text without relying on a fixed subword vocabulary, offering a promising approach to noisy and multilingual inputs.
However, their practical value depends on whether they can be \emph{reliably} adapted to downstream tasks.
To this end, we systematically examine this question through \emph{fine-tuning behavior}, \emph{typographical robustness}, and zero-shot \emph{crosslingual transfer}.
Under our tested settings, pixel-based models require longer training and broader labelled-data coverage than text-based models, while delayed learning and learning-rate-dependent failures make adaptation less reliable.
Our analyses further suggest that task-relevant information is less readily accessible in pretrained pixel representations, placing greater demands on fine-tuning.
Once successfully adapted, however, pixel-based models lose substantially less accuracy under typographical noise and achieve higher target-language accuracy than English-pretrained text-based counterparts on two crosslingual benchmarks, despite weaker clean English performance.
Together, these findings reveal a central trade-off: replacing tokenization with rendering reduces sensitivity to corrupted and unfamiliar inputs, but shifts a central bottleneck to downstream adaptation.\footnote{We release our code at: \url{https://github.com/wastedelement/pixel-model-analysis}.}
\end{abstract}

\section{Introduction}

\begin{figure}[!t]
    \centering
    \setlength{\belowcaptionskip}{-0.1cm}
    \includegraphics[width=0.99\columnwidth]{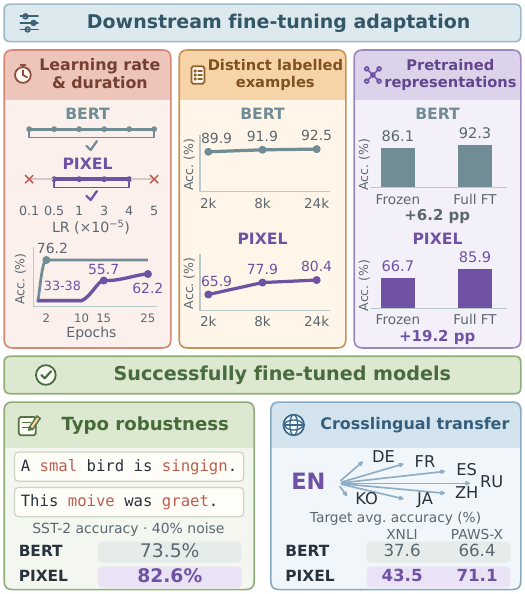}
    \caption{
Overview of the practical challenges and benefits of pixel-based language models.
\textbf{Top}: compared with BERT, PIXEL is more sensitive to learning rate, learns more slowly, benefits more from additional labelled examples, and requires greater adaptation of its pretrained representations.
\textbf{Bottom}: once successfully fine-tuned, PIXEL is more robust to typographical noise and achieves better crosslingual transfer performance.
    }
    \label{fig:teaser}
\end{figure}

Pixel-based language models offer an alternative to subword tokenization by processing text as rendered images \citep{rust2023language,chai-etal-2024-autoregressive,kesen-etal-2025-multilingual}.
Instead of mapping words to entries in a fixed vocabulary, they encode image patches containing the written text \citep{rust2023language}.
This approach is particularly appealing for noisy and multilingual inputs: a typo can break a familiar word into unfamiliar subwords, while languages and scripts poorly covered by a tokenizer may receive 
uninformative representations.
Rendering text avoids dependence on a learned subword vocabulary, motivating research on visual language representations, rendering strategies, and multilingual pixel-based language models \citep{lotz-etal-2023-text,salesky-etal-2023-multilingual,kesen-etal-2025-multilingual}.

The promise of pixel-based language models depends on more than their accuracy on clean benchmarks.
Before their robustness or multilingual coverage can be useful, their pretrained encoders must be reliably adapted to downstream tasks.
\textbf{\emph{Fine-tuning}} is therefore central to evaluating their practical value.
Text-based models such as BERT learn contextual language representations by predicting masked tokens \citep{devlin-etal-2019-bert}, whereas pixel-based encoders reconstruct masked image patches \citep{rust2023language}.
Although visual reconstruction also supports linguistic learning, existing probes suggest that semantic information is less readily accessible in the resulting representations than in BERT's \citep{tatariya-etal-2024-pixology}.
This raises important questions of \emph{how much downstream adaptation pixel-based representations require} and \emph{what allows this adaptation to succeed}.

At the same time, the visual interface may provide distinct advantages once fine-tuning succeeds.
Typographical errors often alter only local visual features while preserving the broader input structure.
Similarly, rendering allows models to process new languages without depending on coverage from a fixed subword vocabulary.
Pixel-based models may therefore be harder to adapt, yet more resilient when applied beyond clean English inputs.
We investigate both sides of this trade-off.

To this end, we systematically compare pixel-based and text-based language models across downstream adaptation, typographical robustness, and zero-shot crosslingual transfer (Figure~\ref{fig:teaser}).
We first examine learning-rate sensitivity, fine-tuning duration, and labelled-data requirements, and then investigate whether pretrained representations contribute to adaptation difficulty.
Finally, we evaluate successfully fine-tuned models under typographical noise and on languages unseen during downstream training.
The key findings are as follows:

\begin{itemize}
    \item \textbf{Pixel-based models require more downstream adaptation} (\secref{fine-tuning}).
    They require more careful optimization, longer training, and broader labelled-data coverage, while task-relevant information appears less readily accessible in their pretrained representations.

    \item \textbf{Pixel-based models show strong resilience to typographical errors} (\secref{typo}).
    Successfully fine-tuned models retain more of their performance as noise increases and can outperform text-based counterparts under heavy noise.

    \item \textbf{Pixel-based models support effective zero-shot crosslingual transfer} (\secref{transfer}).
    After English fine-tuning, they achieve higher average target-language performance than the text-based counterparts on two benchmarks.
\end{itemize}

\section{Related Work}

\subsection{Pixel-Based Language Models}

Most language models tokenize text using fixed subword vocabularies, with approaches like BPE \citep{sennrich2016neural}, WordPiece \citep{devlin-etal-2019-bert,song-etal-2021-fast}, and SentencePiece \citep{kudo-2018-subword,kudo-richardson-2018-sentencepiece}.
Tokenizer-free language models offer an alternative by processing text directly as characters or bytes \citep{clark-etal-2022-canine,xue-etal-2022-byt5}.
Further work reduces computational cost by grouping characters or bytes into variable-length units, using learned boundaries or input-dependent segmentation \citep{godey-etal-2022-manta,nawrot-etal-2023-efficient,pagnoni-etal-2025-byte}.

Pixel-based language models take a visual approach, encoding rendered text as image patches \citep{rust2023language}.
Research in this direction has explored rendering strategies \citep{lotz-etal-2023-text}, autoregressive pretraining \citep{chai-etal-2024-autoregressive,tai-etal-2024-pixar}, and multilingual representations \citep{salesky-etal-2023-multilingual,kesen-etal-2025-multilingual}.
Linguistic probing further examines what these visual representations learn about language \citep{tatariya-etal-2024-pixology}.
Alongside these developments, studies of text-based models have seen substantial variation in fine-tuning outcomes \citep{dodge2020fine,mosbach2021stability,liu2021empirical}.
We bring these questions together by examining how reliably pixel-based models learn downstream tasks, how much supervision they require, and how their pretrained representations support adaptation.

\subsection{Robustness and Crosslingual Transfer}

Typographical robustness concerns whether models preserve their predictions when small changes alter the written form of an input.
Character-level attacks and misspellings can substantially degrade NLP systems, motivating methods for robust text processing \citep{ebrahimi-etal-2018-hotflip,pruthi-etal-2019-combating,moradi-samwald-2021-evaluating}.
Byte-level and visual representations have shown resilience to such perturbations \citep{xue-etal-2022-byt5,salesky-etal-2021-robust,rust2023language}.
Recent work introduces more realistic, language-specific typing errors \citep{zhao-etal-2026-evaluating}.
We build on this work by comparing pixel-based and text-based models under typographical noise, examining both clean accuracy and degradation as noise increases.

Crosslingual transfer enables models to apply knowledge learned in one source language to tasks in other target languages, reducing the need for labelled data in each language \citep{pires-etal-2019-multilingual,dufter-schutze-2020-identifying,conneau-etal-2020-emerging,lauscher-etal-2020-zero,ruder-etal-2021-xtreme,de-vries-etal-2022-make,liu-etal-2023-crosslingual-transfer,liu-etal-2024-ofa}.
Visual text representations have also been shown to boost crosslingual transfer through shared visual patterns, especially for translation and language understanding \citep{salesky-etal-2023-multilingual,kesen-etal-2025-multilingual}.
Building on this work, we examine how well pixel-based models transfer the task knowledge acquired during source-language fine-tuning to other languages.

\section{General Experimental Setup}

\subsection{Models}

\paragraph{Pixel-based models.}
Our primary pixel-based model is \textbf{\pixel{}-base}, pretrained through masked autoencoding on rendered English Wikipedia and BookCorpus \citep{rust2023language}.
We also include \textbf{English \pixel{}-bigrams} and the smaller \textbf{\pixel{}-tiny-continuous} to examine models with different rendering schemes and capacities.
For multilingual pretraining, we include \textbf{\pixel{}-M4}, which uses bigram rendering and is pretrained on English, Hindi, Ukrainian, and Chinese \citep{kesen-etal-2025-multilingual}.

\paragraph{Text-based models.}
Our primary text-based model is \textbf{BERT-base} \citep{devlin-etal-2019-bert}, whose English pretraining sources broadly match those of \pixel{}-base.
This pair forms the main comparison throughout our analyses.
We additionally include \textbf{RoBERTa} as a monolingual reference model \citep{liu2019roberta}, and \textbf{mBERT} \citep{devlin-etal-2019-bert,pires-etal-2019-multilingual} and \textbf{XLM-R} \citep{conneau2020unsupervised} as multilingual reference models.

\subsection{Datasets}

We evaluate models on tasks covering different aspects of language understanding, including sentiment analysis, grammatical acceptability, natural language inference, paraphrase identification, and topic classification.
The datasets are introduced below, 
with more details presented 
in \secref{dataset-details}.

\paragraph{SST-2.}
SST-2 is an English sentiment classification dataset with positive and negative labels for movie-review text \citep{socher2013recursive}.

\paragraph{MNLI.}
MNLI is an English natural language inference dataset containing premise--hypothesis pairs from multiple genres, labelled as entailment, contradiction, or neutral \citep{williams2018broad}.

\paragraph{CoLA.}
CoLA is an English grammatical acceptability classification dataset containing sentences labelled as grammatically acceptable or unacceptable \citep{warstadt2019neural}.

\paragraph{XNLI.}
XNLI is a multilingual natural language inference benchmark with the same three-way classification task as MNLI \citep{conneau2018xnli}.

\paragraph{PAWS-X.}
PAWS-X is a multilingual paraphrase identification dataset containing sentence pairs with high lexical overlap but potentially different meanings \citep{yang-etal-2019-paws-x}.

\paragraph{SIB-200.}
SIB-200 is a multilingual sentence-level topic classification dataset derived from FLORES-200 \citep{nllbteam2022languageleftbehindscaling}.
The dataset is annotated with seven labels \citep{adelani-etal-2024-sib}.

\paragraph{Taxi1500.}
Taxi1500 is a multilingual text classification dataset constructed from the Bible and annotated with six topics \citep{ma-etal-2025-taxi1500}.

\subsection{Fine-Tuning and Evaluation}

We fine-tune pretrained models on labelled English data for each task.
For each task, we initialize the encoder from a pretrained checkpoint, add a classification head, and fine-tune both components simultaneously.\footnote{We also compare training from scratch and training only a classifier on a frozen pretrained encoder in \secref{pretraining}.}
We select checkpoints based on English development performance and compare models on the same hold-out evaluation examples.

Our analyses examine both the adaptation process and the behavior of the fine-tuned models.
For \textbf{fine-tuning analysis}, we study how training conditions and labelled data affect learning and the resulting representations (\secref{fine-tuning}).
For \textbf{typographical robustness}, we evaluate models trained on clean text on both clean inputs and their perturbed counterparts, without additional training on noisy data (\secref{typo}).
For \textbf{zero-shot crosslingual transfer}, we evaluate models fine-tuned on English on target-language inputs without target-language supervision (\secref{transfer}).
Details are described in the corresponding sections.

\begin{figure*}[t]
  \centering
    \setlength{\belowcaptionskip}{-0.3cm}
  \includegraphics[width=\textwidth]{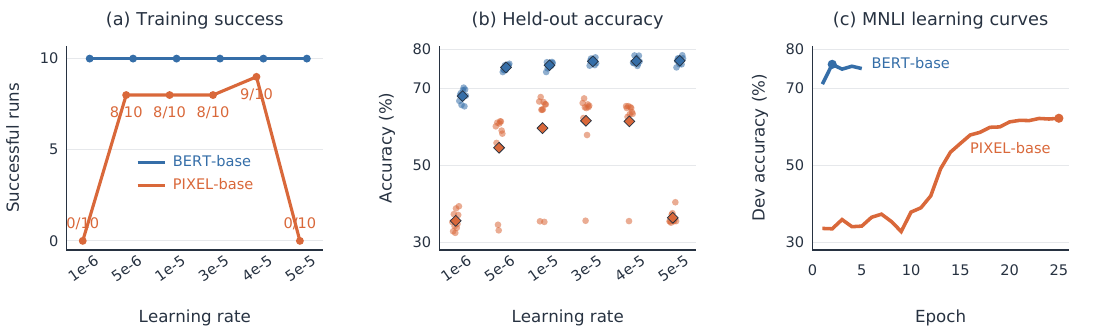}
  \caption{Learning-rate sensitivity and fine-tuning duration. (\textbf{a}) Successful MNLI runs at epoch 10, out of ten seeds per learning rate. (\textbf{b}) Held-out accuracy of development-selected checkpoints from the same runs. Dots show individual runs; diamonds show means across all ten seeds. (\textbf{c}) Development accuracy at each epoch in one recorded MNLI run per model.
  PIXEL is less stable across learning rates and can require longer fine-tuning than BERT.
  }\label{fig:mnli-lr-budget}
\end{figure*}

\section{Fine-Tuning Analysis}\seclabel{fine-tuning}

We examine what makes downstream adaptation challenging for pixel-based language models.
Our analysis addresses three questions: (1) \emph{how optimization settings and training duration affect adaptation}, (2) \emph{how labelled-data coverage affects generalization}, and (3) \emph{whether the pretraining--adaptation gap contributes to fine-tuning instability}.
We primarily compare \textbf{\pixel{}-base} and \textbf{BERT-base} on MNLI 
and SST-2
in this section.\footnote{
Further details and analyses, including renderer--checkpoint compatibility, are provided in \secref{fine-tuning-appendix}.}

\subsection{Optimization and Fine-Tuning Duration}
\seclabel{optimization-range}

We compare the two models across six learning rates on a fixed subset of 24,000 \textbf{MNLI} examples, using ten random seeds per rate.
Both models share the optimization settings and a ten-epoch observation window under the same learning-rate schedule.
We assess training progress and evaluate development-selected checkpoints on separate holdout examples.\footnote{The complete training recipe and per-seed results are given in \secref{training-common-settings} and \secref{lr-complete-results}; schedule follow-ups appear in \secref{trajectory-followup}.}
We additionally plot development accuracy at each epoch in recorded MNLI runs, using each model's own training schedule.
The results are demonstrated in Figure~\ref{fig:mnli-lr-budget}.

\paragraph{Pixel-based models are more sensitive to learning rate.}
Within the same ten-epoch budget, BERT trains successfully across all ten seeds at every tested learning rate.\footnote{We consider a run successful if its training loss is below 1.0 and its training accuracy exceeds 50\% at epoch 10.}
For \pixel{}, only eight or nine of the ten runs succeed at intermediate learning rates, and none succeed at the lowest or highest rates (Figure~\ref{fig:mnli-lr-budget}a).
This sensitivity is particularly clear between $4\times10^{-5}$ and $5\times10^{-5}$: this small increase reduces the number of successful runs from nine to zero and lowers mean held-out accuracy from 61.4\% to 36.4\% (Figure~\ref{fig:mnli-lr-budget}b).
By comparison, BERT succeeds in every run and maintains approximately 77\% accuracy at both rates.
Thus, \pixel{} has a narrower effective learning-rate range than BERT, and even rates within this range do not consistently produce successful training.

\paragraph{Pixel-based models can require longer fine-tuning.}
The development curves show that \pixel{} can begin learning substantially later than BERT (Figure~\ref{fig:mnli-lr-budget}c).
In the illustrated MNLI runs, BERT reaches 76.2\% development accuracy by epoch 2, whereas \pixel{} remains between 33\% and 38\% throughout the first 10 epochs.
\pixel{} begins to improve sharply only after this point, reaching 55.7\% at epoch 15 and 62.2\% at epoch 25.
Continuing beyond the first 10 epochs therefore yields a further 24.3-point improvement over its best accuracy within that initial window.
This pattern also appears across multiple runs in our longer-budget analysis reported in Appendix~\ref{sec:historical-seeds}: among the seven seeds for which \pixel{} trains successfully, its best checkpoint occurs at a median of 21 epochs, compared with only 3 epochs for BERT on the same seeds.
\textbf{A short fine-tuning schedule may therefore stop \pixel{} before substantial learning begins, making a longer training window and greater patience necessary.}

\subsection{Supervision Budget}

\begin{figure}[ht]
  \centering
    \setlength{\belowcaptionskip}{-0.3cm}
  \includegraphics[width=\columnwidth]{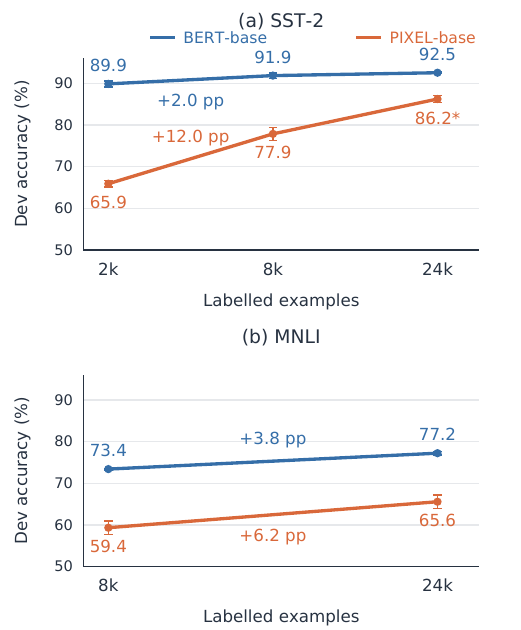}
  \caption{Development accuracy with increasing numbers of distinct labelled examples under a fixed budget of 15,000 updates.
Lines show means and error bars show standard deviations across three seeds. 
Individual scores appear in Table~\ref{tab:sample-budget-seeds}.
Across both tasks, BERT performs strongly with fewer distinct examples, whereas \pixel{} benefits more from broader data coverage.}
  \label{fig:sample-coverage}
  \end{figure}

We examine \emph{how the amount of labelled data affects downstream performance} while keeping the number of training updates fixed.
We train both models on SST-2 subsets containing 2,000, 8,000, or 24,000 examples, and MNLI subsets containing 8,000 or 24,000 examples.
Every run uses 15,000 updates with an effective batch size of 128.
Each run therefore processes approximately the same total number of examples, but smaller training subsets are repeated more often.
For the cross-budget comparison, we use three seeds per setting and report the best recorded development accuracy.
\footnote{Subsets, model-specific learning rates, and complete results are detailed in Appendix~\ref{sec:coverage-details}.}
The results are demonstrated in Figure~\ref{fig:sample-coverage}.

\paragraph{Pixel-based models benefit more from additional labelled examples.}
Under the same update budget, BERT gains only modestly from seeing more distinct training examples.
Increasing the SST-2 training set from 2,000 to 8,000 examples improves its development accuracy by 2.0 points (Figure~\ref{fig:sample-coverage}a), while increasing the MNLI training set from 8,000 to 24,000 examples produces a 3.8-point gain (Figure~\ref{fig:sample-coverage}b).
In comparison, \pixel{} gains 12.0 points on SST-2 and 6.2 points on MNLI from the corresponding increases in labelled data.
Because the number of training updates remains fixed (15,000 updates), these gains should be attributed to greater coverage of distinct labelled examples instead of additional optimization.
Moreover, all runs finish with more than 99.5\% online training accuracy, including those trained on the smallest subsets.
Both models can therefore fit the available training data, but \pixel{} relies more heavily on broader labelled-data coverage to generalize beyond it.\footnote{Additional SST-2 experiments show that greater labelled data coverage does not eliminate variation and poor generalization across \pixel{} runs. 
In a ten-seed experiment with 24,000 examples, all runs exceed 99.8\% training accuracy, but two reach only 68.1--68.6\% development accuracy, compared with 85.7--88.0\% for the remaining runs; their test performance is similarly low. See Appendix~\ref{sec:coverage-details} for complete results.}

\begin{figure*}[t]
\centering
\setlength{\abovecaptionskip}{-0.0001cm}
\includegraphics[width=\textwidth]{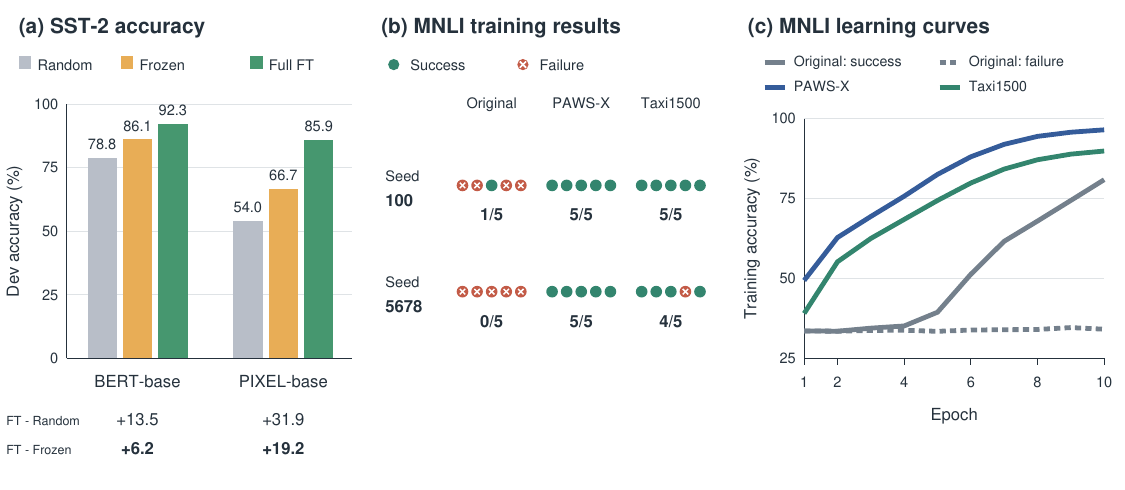}
\caption{Pretraining and downstream adaptation.
(\textbf{a}) SST-2 development accuracy when training from random initialization (\texttt{Random}), training a classifier on a frozen pretrained encoder (\texttt{Frozen}), or fine-tuning the full pretrained model (\texttt{Full FT}).
(\textbf{b}) MNLI training success across five repeated runs for two seeds, comparing the original \pixel{} checkpoint with checkpoints previously adapted on PAWS-X or Taxi1500.
(\textbf{c}) Mean MNLI learning curves; original runs are split by outcome.
\pixel{} benefits substantially from adapting its pretrained representations, while prior supervised downstream task training makes subsequent fine-tuning faster and more reliable.}
\label{fig:pretraining-gap}
\end{figure*}

\subsection{Pretraining--Adaptation Gap}\seclabel{pretraining}

The preceding experiments show that \pixel{} is harder to fine-tune than BERT and can fail under settings in which BERT trains reliably.
This raises a further question: \emph{does pixel-based pretraining produce representations that require more adaptation before they can support downstream language tasks}?
We investigate this question through two complementary experiments.

First, we examine how much downstream performance is already supported by the pretrained representations (Figure~\ref{fig:pretraining-gap}a).
We compare BERT-base and \pixel{}-base on SST-2 under three different training conditions.
In \texttt{Random}, the encoder and classifier are initialized randomly and trained jointly.
In \texttt{Frozen}, the pretrained encoder remains fixed and only a linear classifier is trained on its mean-pooled representations.
In \texttt{Full FT}, both the pretrained encoder and classifier are trained jointly.
The comparison between \texttt{Random} and \texttt{Full FT} measures the value of pretrained initialization, while the comparison between \texttt{Frozen} and \texttt{Full FT} indicates how much the model benefits from adapting its pretrained representations to the task.

Second, we test whether previous supervised adaptation to another task makes subsequent MNLI fine-tuning more reliable (Figure~\ref{fig:pretraining-gap}b--c).
Instead of initializing \pixel{} from its original pretrained checkpoint, we use checkpoints previously fine-tuned on PAWS-X paraphrase identification or Taxi1500 topic classification.
Neither task directly teaches MNLI-style inference, but both provide additional supervised training on sentence-level language understanding.
If these checkpoints improve MNLI training, this would suggest that prior task adaptation provides \pixel{} with a more suitable starting point for learning a new downstream task.

\paragraph{PIXEL requires substantially more downstream adaptation to become task-ready.}
Pretraining benefits both models, but the gap between frozen representations and full fine-tuning is much larger for \pixel{}.
On SST-2, \pixel{} reaches 54.0\% accuracy from random initialization and 85.9\% with full fine-tuning, showing a 31.9-point benefit from pretrained initialization (Figure~\ref{fig:pretraining-gap}a).
However, training only a classifier on the frozen encoder reaches 66.7\%, leaving a further 19.2-point improvement from adapting the encoder.
For BERT, the corresponding accuracies are 78.8\%, 86.1\%, and 92.3\% under \texttt{Random}, \texttt{Frozen}, and \texttt{Full FT}, respectively.
Its frozen representations are therefore only 6.2 points behind full fine-tuning.
While BERT's pretrained representations already provide a strong basis for SST-2 classification, much of \pixel{}'s downstream performance emerges only after its encoder is adapted.
This suggests that task-relevant information is less readily accessible in pretrained pixel representations.

\paragraph{Prior language-task adaptation provides PIXEL with a more reliable starting point.}
We revisit two MNLI seeds that previously showed unstable training and rerun each seed five times from the original \pixel{} checkpoint and from checkpoints adapted on PAWS-X or Taxi1500.
For seed 100, the original checkpoint succeeds in only one of five runs, whereas both adapted checkpoints succeed in all five.
For seed 5678, all five runs from the original checkpoint fail, while PAWS-X succeeds in all five runs and Taxi1500 in four (Figure~\ref{fig:pretraining-gap}b).
The adapted checkpoints also begin learning earlier; by contrast, successful runs from the original checkpoint improve only after an initial plateau, while failed runs remain near chance accuracy (Figure~\ref{fig:pretraining-gap}c).
Because PAWS-X and Taxi1500 involve different classification tasks, and their task-specific classifiers are discarded before MNLI fine-tuning, the improvement should be attributed to the change of encoders.
The results therefore identify the pretrained encoder representation as an important contributor to \pixel{}'s fine-tuning instability.

\section{Typographical Robustness}\seclabel{typo}

The previous analysis examined the challenges of adapting pixel-based models to downstream tasks.
We now ask whether, once successfully fine-tuned, these models are more robust to typographical errors (typos).
Typos are common in user-written text \citep{conijn2019typo,kumar-etal-2020-noisy} and can change how models tokenize and represent otherwise familiar words \citep{chai-etal-2024-tokenization}.
We examine both \emph{how typos affect input length} and \emph{how well models maintain their task performance}.

\subsection{Experimental Setup}

\begin{figure*}[t]
  \centering
    \setlength{\abovecaptionskip}{-0.0001cm}
  \includegraphics[width=\textwidth]{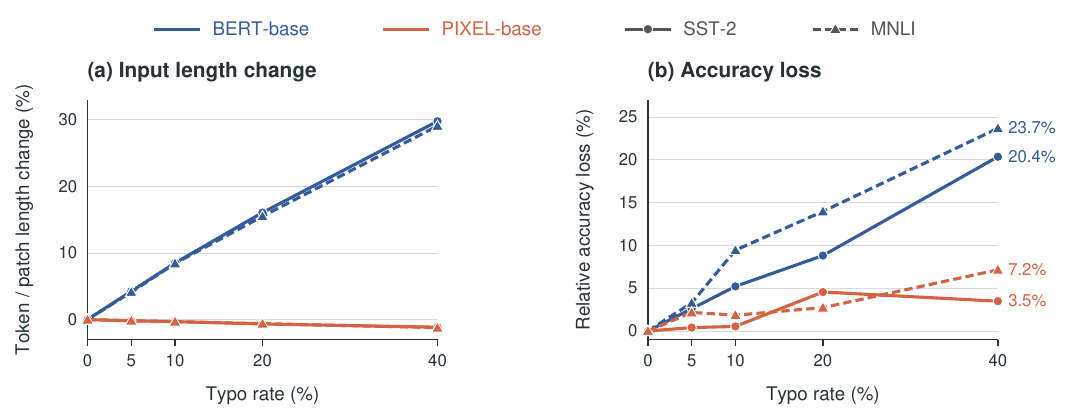}
  \caption{Effects of typographical noise on input length and task performance.
  (\textbf{a}) Percentage change in mean input length relative to clean inputs, measured in subword tokens for BERT and image patches for \pixel{}. (\textbf{b}) Accuracy loss as a percentage of each model's clean accuracy. Colors distinguish BERT-base and \pixel{}-base; solid lines with circles show SST-2, and dashed lines with triangles show MNLI. Complete model scores appear in Table~\ref{tab:english-robustness}.
    }
  \label{fig:robustness}
\end{figure*}

We evaluate typographical robustness on SST-2 sentiment classification and MNLI natural language inference \citep{socher2013recursive,williams2018broad}.
Our primary comparison is between BERT-base and \pixel{}-base.
For each task, every model is fine-tuned on a fixed subset of 24,000 clean training examples, and checkpoints are selected based on clean development performance.
The selected checkpoints are then evaluated, without further training, on clean evaluation examples and their typo-perturbed counterparts.\footnote{Complete training and evaluation details, together with results for all models, are provided in \secref{robustness-appendix}. Additional CoLA fine-tuning results are reported in \secref{additional-finetuning}.}

We generate the perturbed inputs using \textsc{MulTypo}, which simulates common typing mistakes through text-editing and keyboard-related operations \citep{zhao-etal-2026-evaluating}.
We apply typo rates of 5\%, 10\%, 20\%, and 40\%, retain the original task labels, and present the same perturbed examples to every model.
We measure performance degradation as the percentage of each model's clean accuracy that is lost under noise.
To examine how typos affect the input interface, we also report the percentage change in mean input length relative to clean inputs, measured in subword tokens for text-based models and image patches for pixel-based models.
Figure~\ref{fig:robustness} presents input-length change and relative accuracy loss for BERT-base and \pixel{}-base, while Table~\ref{tab:english-robustness} in the appendix reports absolute accuracy for all evaluated models.

\subsection{Results and Discussion}

\paragraph{Typos expand BERT's token sequences but leave PIXEL's patch sequences nearly unchanged.}
BERT's mean input length increases steadily with the typo rate, reaching 29.8\% on SST-2 and 29.1\% on MNLI at 40\% noise (Figure~\ref{fig:robustness}a).
By contrast, \pixel{}-base's mean patch length decreases by approximately 1\% on both tasks, which is consistent with deletion operations slightly shortening the rendered text.
These contrasting responses reflect how the two input interfaces handle misspellings.
A typo can split a familiar word into multiple subword tokens, changing both the identities and positions of subsequent tokens.
For \pixel{}, the same typo changes local visual features but generally preserves the overall number of patches.
The rendered interface therefore avoids the substantial sequence expansion observed with subword tokenization.

\paragraph{PIXEL-base retains more of its clean accuracy under typographical noise.}
Across both tasks, BERT's relative accuracy loss increases much more sharply than that of \pixel{}-base (Figure~\ref{fig:robustness}b).
At 40\% noise, BERT loses 20.4\% of its clean accuracy on SST-2 and 23.7\% on MNLI, whereas \pixel{} loses only 3.5\% and 7.2\%, respectively.
This consistent gap shows that \pixel{} is less sensitive to typographical perturbations despite receiving altered visual inputs.
Together with the input-length results, the robustness appears to reflect two properties: avoiding typo-induced sequence fragmentation and retaining the ability to map altered visual features to useful linguistic representations.

\begin{table*}[t]
  \centering
  \fontsize{8}{9.5}\selectfont
  \setlength{\tabcolsep}{2.4pt}
  \renewcommand{\arraystretch}{0.88}
  \begin{tabular*}{\textwidth}{@{\extracolsep{\fill}}l*{7}{r}@{\hspace{9pt}}*{8}{r}@{}}
    \toprule
    & \multicolumn{7}{c}{\textbf{XNLI}}
    & \multicolumn{8}{c}{\textbf{PAWS-X}} \\
    \cmidrule(lr){2-8}\cmidrule(lr){9-16}
    \textbf{Model}
    & EN & DE & FR & ES & RU & ZH & Avg.
    & EN & DE & FR & ES & ZH & JA & KO & Avg. \\
    \midrule
    BERT-base
    & \textbf{76.8} & 38.9 & 38.7 & 40.5 & 34.1 & \textbf{35.8} & 37.6
    & \textbf{93.6} & 75.3 & 79.0 & 76.8 & 52.9 & 57.1 & 57.4 & 66.4 \\
    \pixel{}-base
    & 65.7 & \textbf{43.2} & \textbf{50.2} & \textbf{49.1} & \textbf{41.4} & 33.5 & \textbf{43.5}
    & 89.1 & \textbf{77.2} & \textbf{80.8} & \textbf{80.5} & \textbf{64.7} & \textbf{61.6} & \textbf{61.9} & \textbf{71.1} \\
    \midrule
    RoBERTa
    & 83.2 & 51.0 & 59.8 & 61.9 & 39.0 & 41.2 & 50.6
    & 94.9 & 83.4 & 83.9 & 84.7 & 66.8 & 63.1 & 61.6 & 73.9 \\
    \pixel{}-bigrams
    & 60.6 & 44.9 & 47.4 & 48.6 & 42.4 & 34.1 & 43.5
    & 89.4 & 77.9 & 79.8 & 80.9 & 57.1 & 56.8 & 61.6 & 69.0 \\
    \addlinespace[2pt]
    mBERT
    & 74.2 & 69.1 & 68.7 & 68.3 & 65.2 & 66.2 & 67.5
    & 94.0 & 86.1 & 87.2 & 86.6 & 76.8 & 74.5 & 71.0 & 80.4 \\
    XLM-R
    & 77.7 & 72.4 & 72.0 & 73.4 & 71.9 & 69.7 & 71.9
    & 94.7 & 87.4 & 89.9 & 89.1 & 81.5 & 77.1 & 76.1 & 83.5 \\
    \pixel{}-M4
    & 62.1 & 44.7 & 45.2 & 47.2 & 45.8 & 40.4 & 44.7
    & 89.3 & 71.6 & 76.6 & 74.2 & 68.9 & 63.1 & 64.5 & 69.8 \\
    \bottomrule
  \end{tabular*}
  \caption{Zero-shot crosslingual transfer performance (\%).
  Avg.\ is the average over target languages.
  \textbf{Bold} compares BERT-base and \pixel{}-base within each column; other models provide reference results.
  Despite lower English accuracy, \pixel{}-base achieves higher target-language averages than BERT-base on both tasks.}
  \label{tab:transfer-main}
\end{table*}

\section{Zero-Shot Cross-Lingual Transfer}\seclabel{transfer}

After examining robustness to noisy English inputs, we turn to a different form of generalization: transferring a task learned in English to other languages.
We ask \emph{whether pixel-based models transfer more effectively than text-based models} and \emph{whether higher English accuracy necessarily leads to stronger crosslingual performance}.

\subsection{Experimental Setup}

We fine-tune each model on English training data, select its checkpoint solely by English development performance, and evaluate it directly on target-language test sets.
Our primary comparison is between BERT-base and \pixel{}-base, whose English pretraining data are broadly comparable.
We additionally include RoBERTa, \pixel{}-bigrams, \pixel{}-M4, mBERT, and XLM-R as reference models with different pretraining data, language coverage, and training objectives.

We focus on XNLI and PAWS-X.
For XNLI, models are trained on 20,000 English MNLI examples and evaluated on German, French, Spanish, Russian, and Chinese \citep{conneau2018xnli}.
For PAWS-X, models are trained on 49,401 English paraphrase pairs and evaluated on German, French, Spanish, Chinese, Japanese, and Korean \citep{yang-etal-2019-paws-x}.
We report English accuracy and accuracy in each target language in Table~\ref{tab:transfer-main}.\footnote{Additional SIB-200 \citep{adelani-etal-2024-sib} and Taxi1500 \citep{ma-etal-2025-taxi1500} results are reported in \secref{transfer-appendix}.}

\subsection{Results and Discussion}

\paragraph{PIXEL-base transfers more effectively than BERT-base despite weaker English performance.}
BERT achieves substantially higher English accuracy than \pixel{}-base on both XNLI and PAWS-X, but this advantage does not carry over to the target languages.
Without receiving any target-language supervision, \pixel{} obtains higher target-language averages on both tasks and outperforms BERT in four of the five XNLI languages and all six PAWS-X languages (Table~\ref{tab:transfer-main}).
The advantage is particularly clear for languages that share the Latin script with English: on XNLI, \pixel{} exceeds BERT by 11.5 points in French and 8.6 points in Spanish.
It also extends to some languages using different scripts, including Russian on XNLI and Chinese, Japanese, and Korean on PAWS-X.
We attribute this broader transfer partly to the absence of a fixed subword vocabulary.
BERT may fragment unfamiliar target-language words into subword sequences poorly matched to those seen during English fine-tuning, whereas \pixel{} can reuse its adapted visual representations without depending on target-language vocabulary coverage.

\paragraph{Multilingual pretraining provides the strongest basis for crosslingual transfer.}
mBERT and XLM-R achieve the highest target-language averages on both benchmarks.
This is expected because large-scale multilingual pretraining improves zero-shot transfer and that stronger crosslingual representation alignment is associated with better target-language performance \citep{liu-etal-2024-translico,liu-etal-2025-langsamp}.
Exposure to multiple languages during pretraining can place their representations in a more language-neutral space, allowing the task-specific capability learned through English fine-tuning to operate more effectively on target-language inputs.
However, multilingual exposure alone seems to be insufficient.
Although \pixel{}-M4 includes Chinese during pretraining, it remains far behind mBERT and XLM-R on Chinese in both tasks.
We attribute this gap partly to weaker adaptation: \pixel{}-M4 already performs considerably worse than its multilingual counterparts on English, leaving it with a less effective task-specific capability to transfer.

\section{Conclusion}

We systematically examined pixel-based language models through downstream fine-tuning, typographical robustness, and zero-shot crosslingual transfer. 
Our results identify downstream adaptation as their central practical challenge: pixel-based models require more careful optimization, longer training, and broader labelled-data coverage, suggesting that task-relevant information is less readily accessible in their pretrained representations. 
Once successfully adapted, however, they show greater resilience to typographical noise and stronger transfer to unseen languages than English-pretrained text-based counterparts. 
These findings reveal a central trade-off: removing a fixed subword vocabulary reduces sensitivity to corrupted and unfamiliar inputs, but places greater demands on downstream adaptation. 
The practical value of pixel-based models should therefore be assessed through both adaptation reliability and generalization beyond clean source-language inputs.

\section*{Limitations}

We provide a systematic evaluation of downstream adaptation, typographical robustness, and zero-shot crosslingual transfer in pixel-based models. However, we acknowledge three main limitations.

First, due to computational constraints, our experiments focus on encoder-based models and language understanding tasks. 
Future work could extend this evaluation to autoregressive pixel-based models and generative tasks.

Second, the evaluated pixel- and text-based models differ in architecture, pretraining data, and training objectives. Our findings therefore characterize the behavior of the tested models but do not fully isolate the causes of their differences. 
Controlled comparisons and mechanistic interpretability analyses could provide more direct explanations, which we leave for future work.

Finally, our fine-tuning experiments focus on learning rate, training duration, labelled-data coverage, and pretrained initialization.
Other choices, including weight decay, optimization algorithms, and parameter-efficient methods such as LoRA \citep{hu2022iclr-lora}, may also affect adaptation. 
An exhaustive evaluation of all these dimensions is beyond our computational budget and can be explored in future work.

\section*{Ethical Considerations}

\paragraph{Use of AI Assistants.}
The authors used an AI assistant for manuscript drafting and language editing, code and figure development, and exploratory organization of analyses.\footnote{\url{https://chatgpt.com/}}
All reported experimental results and scientific interpretations were reviewed by the authors, who take responsibility for the final paper.

\section*{Acknowledgments}

This research was supported by the Munich Center
for Machine Learning (MCML) and German Research Foundation (DFG, grant SCHU 2246/14-1).

\FloatBarrier \bibliography{references} \clearpage

\begin{thebibliography}{49}
\providecommand{\natexlab}[1]{#1}

\bibitem[{Adelani et~al.(2024)Adelani, Liu, Shen, Vassilyev, Alabi, Mao, Gao,
  and Lee}]{adelani-etal-2024-sib}
David~Ifeoluwa Adelani, Hannah Liu, Xiaoyu Shen, Nikita Vassilyev, Jesujoba~O.
  Alabi, Yanke Mao, Haonan Gao, and En-Shiun~Annie Lee. 2024.
\newblock \href {https://doi.org/10.18653/v1/2024.eacl-long.14} {{SIB}-200: A
  simple, inclusive, and big evaluation dataset for topic classification in
  200+ languages and dialects}.
\newblock In \emph{Proceedings of the 18th Conference of the European Chapter
  of the Association for Computational Linguistics (Volume 1: Long Papers)},
  pages 226--245, St. Julian{'}s, Malta. Association for Computational
  Linguistics.

\bibitem[{Chai et~al.(2024{\natexlab{a}})Chai, Fang, Peng, and
  Li}]{chai-etal-2024-tokenization}
Yekun Chai, Yewei Fang, Qiwei Peng, and Xuhong Li. 2024{\natexlab{a}}.
\newblock \href {https://doi.org/10.18653/v1/2024.findings-emnlp.86}
  {Tokenization falling short: On subword robustness in large language models}.
\newblock In \emph{Findings of the Association for Computational Linguistics:
  EMNLP 2024}, pages 1582--1599, Miami, Florida, USA. Association for
  Computational Linguistics.

\bibitem[{Chai et~al.(2024{\natexlab{b}})Chai, Liu, Xiao, Wang, Sun, and
  Wu}]{chai-etal-2024-autoregressive}
Yekun Chai, Qingyi Liu, Jingwu Xiao, Shuohuan Wang, Yu~Sun, and Hua Wu.
  2024{\natexlab{b}}.
\newblock \href {https://doi.org/10.18653/v1/2024.emnlp-main.182}
  {Autoregressive pre-training on pixels and texts}.
\newblock In \emph{Proceedings of the 2024 Conference on Empirical Methods in
  Natural Language Processing}, pages 3106--3125, Miami, Florida, USA.
  Association for Computational Linguistics.

\bibitem[{Clark et~al.(2022)Clark, Garrette, Turc, and
  Wieting}]{clark-etal-2022-canine}
Jonathan~H. Clark, Dan Garrette, Iulia Turc, and John Wieting. 2022.
\newblock \href {https://doi.org/10.1162/tacl_a_00448} {Canine: Pre-training an
  efficient tokenization-free encoder for language representation}.
\newblock \emph{Transactions of the Association for Computational Linguistics},
  10:73--91.

\bibitem[{Conijn et~al.(2019)Conijn, Van~Zaanen, Leijten, and
  Van~Waes}]{conijn2019typo}
Rianne Conijn, Menno Van~Zaanen, Mari{\"e}lle Leijten, and Luuk Van~Waes. 2019.
\newblock \href {https://doi.org/10.37514/JWA-J.2019.3.1.05} {How to typo?
  building a process-based model of typographic error revisions}.
\newblock \emph{Journal of Writing Analytics}, 3:69--95.

\bibitem[{Conneau et~al.(2020{\natexlab{a}})Conneau, Khandelwal, Goyal,
  Chaudhary, Wenzek, Guzm{\'a}n, Grave, Ott, Zettlemoyer, and
  Stoyanov}]{conneau2020unsupervised}
Alexis Conneau, Kartikay Khandelwal, Naman Goyal, Vishrav Chaudhary, Guillaume
  Wenzek, Francisco Guzm{\'a}n, Edouard Grave, Myle Ott, Luke Zettlemoyer, and
  Veselin Stoyanov. 2020{\natexlab{a}}.
\newblock \href {https://doi.org/10.18653/v1/2020.acl-main.747} {Unsupervised
  cross-lingual representation learning at scale}.
\newblock In \emph{Proceedings of the 58th Annual Meeting of the Association
  for Computational Linguistics}, pages 8440--8451, Online. Association for
  Computational Linguistics.

\bibitem[{Conneau et~al.(2018)Conneau, Rinott, Lample, Williams, Bowman,
  Schwenk, and Stoyanov}]{conneau2018xnli}
Alexis Conneau, Ruty Rinott, Guillaume Lample, Adina Williams, Samuel~R.
  Bowman, Holger Schwenk, and Veselin Stoyanov. 2018.
\newblock \href {https://doi.org/10.18653/v1/D18-1269} {{XNLI}: Evaluating
  cross-lingual sentence representations}.
\newblock In \emph{Proceedings of the 2018 Conference on Empirical Methods in
  Natural Language Processing}, pages 2475--2485, Brussels, Belgium.
  Association for Computational Linguistics.

\bibitem[{Conneau et~al.(2020{\natexlab{b}})Conneau, Wu, Li, Zettlemoyer, and
  Stoyanov}]{conneau-etal-2020-emerging}
Alexis Conneau, Shijie Wu, Haoran Li, Luke Zettlemoyer, and Veselin Stoyanov.
  2020{\natexlab{b}}.
\newblock \href {https://doi.org/10.18653/v1/2020.acl-main.536} {Emerging
  cross-lingual structure in pretrained language models}.
\newblock In \emph{Proceedings of the 58th Annual Meeting of the Association
  for Computational Linguistics}, pages 6022--6034, Online. Association for
  Computational Linguistics.

\bibitem[{de~Vries et~al.(2022)de~Vries, Wieling, and
  Nissim}]{de-vries-etal-2022-make}
Wietse de~Vries, Martijn Wieling, and Malvina Nissim. 2022.
\newblock \href {https://doi.org/10.18653/v1/2022.acl-long.529} {Make the best
  of cross-lingual transfer: Evidence from {POS} tagging with over 100
  languages}.
\newblock In \emph{Proceedings of the 60th Annual Meeting of the Association
  for Computational Linguistics (Volume 1: Long Papers)}, pages 7676--7685,
  Dublin, Ireland. Association for Computational Linguistics.

\bibitem[{Devlin et~al.(2019)Devlin, Chang, Lee, and
  Toutanova}]{devlin-etal-2019-bert}
Jacob Devlin, Ming-Wei Chang, Kenton Lee, and Kristina Toutanova. 2019.
\newblock \href {https://doi.org/10.18653/v1/N19-1423} {{BERT}: Pre-training of
  deep bidirectional transformers for language understanding}.
\newblock In \emph{Proceedings of the 2019 Conference of the North {A}merican
  Chapter of the Association for Computational Linguistics: Human Language
  Technologies, Volume 1 (Long and Short Papers)}, pages 4171--4186,
  Minneapolis, Minnesota. Association for Computational Linguistics.

\bibitem[{Dodge et~al.(2020)Dodge, Ilharco, Schwartz, Farhadi, Hajishirzi, and
  Smith}]{dodge2020fine}
Jesse Dodge, Gabriel Ilharco, Roy Schwartz, Ali Farhadi, Hannaneh Hajishirzi,
  and Noah Smith. 2020.
\newblock \href {https://arxiv.org/abs/2002.06305} {Fine-tuning pretrained
  language models: Weight initializations, data orders, and early stopping}.
\newblock \emph{Preprint}, arXiv:2002.06305.

\bibitem[{Dufter and Sch{\"u}tze(2020)}]{dufter-schutze-2020-identifying}
Philipp Dufter and Hinrich Sch{\"u}tze. 2020.
\newblock \href {https://doi.org/10.18653/v1/2020.emnlp-main.358} {Identifying
  elements essential for {BERT}{'}s multilinguality}.
\newblock In \emph{Proceedings of the 2020 Conference on Empirical Methods in
  Natural Language Processing (EMNLP)}, pages 4423--4437, Online. Association
  for Computational Linguistics.

\bibitem[{Ebrahimi et~al.(2018)Ebrahimi, Rao, Lowd, and
  Dou}]{ebrahimi-etal-2018-hotflip}
Javid Ebrahimi, Anyi Rao, Daniel Lowd, and Dejing Dou. 2018.
\newblock \href {https://doi.org/10.18653/v1/P18-2006} {{H}ot{F}lip: White-box
  adversarial examples for text classification}.
\newblock In \emph{Proceedings of the 56th Annual Meeting of the Association
  for Computational Linguistics (Volume 2: Short Papers)}, pages 31--36,
  Melbourne, Australia. Association for Computational Linguistics.

\bibitem[{Godey et~al.(2022)Godey, Castagn{\'e}, de~la Clergerie, and
  Sagot}]{godey-etal-2022-manta}
Nathan Godey, Roman Castagn{\'e}, {\'E}ric de~la Clergerie, and Beno{\^i}t
  Sagot. 2022.
\newblock \href {https://doi.org/10.18653/v1/2022.findings-emnlp.207} {{MANT}a:
  Efficient gradient-based tokenization for end-to-end robust language
  modeling}.
\newblock In \emph{Findings of the Association for Computational Linguistics:
  EMNLP 2022}, pages 2859--2870, Abu Dhabi, United Arab Emirates. Association
  for Computational Linguistics.

\bibitem[{Hu et~al.(2022)Hu, Shen, Wallis, Allen{-}Zhu, Li, Wang, Wang, and
  Chen}]{hu2022iclr-lora}
Edward~J. Hu, Yelong Shen, Phillip Wallis, Zeyuan Allen{-}Zhu, Yuanzhi Li,
  Shean Wang, Lu~Wang, and Weizhu Chen. 2022.
\newblock \href {https://openreview.net/forum?id=nZeVKeeFYf9} {Lora: Low-rank
  adaptation of large language models}.
\newblock In \emph{The Tenth International Conference on Learning
  Representations, {ICLR} 2022, Virtual Event, April 25-29, 2022}.
  OpenReview.net.

\bibitem[{Kesen et~al.(2025)Kesen, Lotz, Ziegler, Rust, and
  Elliott}]{kesen-etal-2025-multilingual}
Ilker Kesen, Jonas~F. Lotz, Ingo Ziegler, Phillip Rust, and Desmond Elliott.
  2025.
\newblock \href {https://doi.org/10.18653/v1/2025.emnlp-main.1504}
  {Multilingual pretraining for pixel language models}.
\newblock In \emph{Proceedings of the 2025 Conference on Empirical Methods in
  Natural Language Processing}, pages 29594--29611, Suzhou, China. Association
  for Computational Linguistics.

\bibitem[{Kudo(2018)}]{kudo-2018-subword}
Taku Kudo. 2018.
\newblock \href {https://doi.org/10.18653/v1/P18-1007} {Subword regularization:
  Improving neural network translation models with multiple subword
  candidates}.
\newblock In \emph{Proceedings of the 56th Annual Meeting of the Association
  for Computational Linguistics (Volume 1: Long Papers)}, pages 66--75,
  Melbourne, Australia. Association for Computational Linguistics.

\bibitem[{Kudo and Richardson(2018)}]{kudo-richardson-2018-sentencepiece}
Taku Kudo and John Richardson. 2018.
\newblock \href {https://doi.org/10.18653/v1/D18-2012} {{S}entence{P}iece: A
  simple and language independent subword tokenizer and detokenizer for neural
  text processing}.
\newblock In \emph{Proceedings of the 2018 Conference on Empirical Methods in
  Natural Language Processing: System Demonstrations}, pages 66--71, Brussels,
  Belgium. Association for Computational Linguistics.

\bibitem[{Kumar et~al.(2020)Kumar, Makhija, and Gupta}]{kumar-etal-2020-noisy}
Ankit Kumar, Piyush Makhija, and Anuj Gupta. 2020.
\newblock \href {https://doi.org/10.18653/v1/2020.wnut-1.3} {Noisy text data:
  Achilles' heel of {BERT}}.
\newblock In \emph{Proceedings of the Sixth Workshop on Noisy User-generated
  Text (W-NUT 2020)}, pages 16--21, Online. Association for Computational
  Linguistics.

\bibitem[{Lauscher et~al.(2020)Lauscher, Ravishankar, Vuli{\'c}, and
  Glava{\v{s}}}]{lauscher-etal-2020-zero}
Anne Lauscher, Vinit Ravishankar, Ivan Vuli{\'c}, and Goran Glava{\v{s}}. 2020.
\newblock \href {https://doi.org/10.18653/v1/2020.emnlp-main.363} {From zero to
  hero: {O}n the limitations of zero-shot language transfer with multilingual
  {T}ransformers}.
\newblock In \emph{Proceedings of the 2020 Conference on Empirical Methods in
  Natural Language Processing (EMNLP)}, pages 4483--4499, Online. Association
  for Computational Linguistics.

\bibitem[{Liu and Wang(2021)}]{liu2021empirical}
Xueqing Liu and Chi Wang. 2021.
\newblock \href {https://doi.org/10.18653/v1/2021.acl-long.178} {An empirical
  study on hyperparameter optimization for fine-tuning pre-trained language
  models}.
\newblock In \emph{Proceedings of the 59th Annual Meeting of the Association
  for Computational Linguistics and the 11th International Joint Conference on
  Natural Language Processing (Volume 1: Long Papers)}, pages 2286--2300,
  Online. Association for Computational Linguistics.

\bibitem[{Liu et~al.(2024{\natexlab{a}})Liu, Lin, Wang, and
  Schuetze}]{liu-etal-2024-ofa}
Yihong Liu, Peiqin Lin, Mingyang Wang, and Hinrich Schuetze.
  2024{\natexlab{a}}.
\newblock \href {https://doi.org/10.18653/v1/2024.findings-naacl.68} {{OFA}: A
  framework of initializing unseen subword embeddings for efficient large-scale
  multilingual continued pretraining}.
\newblock In \emph{Findings of the Association for Computational Linguistics:
  NAACL 2024}, pages 1067--1097, Mexico City, Mexico. Association for
  Computational Linguistics.

\bibitem[{Liu et~al.(2024{\natexlab{b}})Liu, Ma, Ye, and
  Schuetze}]{liu-etal-2024-translico}
Yihong Liu, Chunlan Ma, Haotian Ye, and Hinrich Schuetze. 2024{\natexlab{b}}.
\newblock \href {https://doi.org/10.18653/v1/2024.acl-long.136} {{T}ransli{C}o:
  A contrastive learning framework to address the script barrier in
  multilingual pretrained language models}.
\newblock In \emph{Proceedings of the 62nd Annual Meeting of the Association
  for Computational Linguistics (Volume 1: Long Papers)}, pages 2476--2499,
  Bangkok, Thailand. Association for Computational Linguistics.

\bibitem[{Liu et~al.(2025)Liu, Ye, Ma, Wang, and
  Schuetze}]{liu-etal-2025-langsamp}
Yihong Liu, Haotian Ye, Chunlan Ma, Mingyang Wang, and Hinrich Schuetze. 2025.
\newblock \href {https://doi.org/10.18653/v1/2025.acl-long.88} {{L}ang{SAMP}:
  Language-script aware multilingual pretraining}.
\newblock In \emph{Proceedings of the 63rd Annual Meeting of the Association
  for Computational Linguistics (Volume 1: Long Papers)}, pages 1743--1770,
  Vienna, Austria. Association for Computational Linguistics.

\bibitem[{Liu et~al.(2023)Liu, Ye, Weissweiler, Pei, and
  Schuetze}]{liu-etal-2023-crosslingual-transfer}
Yihong Liu, Haotian Ye, Leonie Weissweiler, Renhao Pei, and Hinrich Schuetze.
  2023.
\newblock \href {https://doi.org/10.18653/v1/2023.findings-emnlp.562}
  {Crosslingual transfer learning for low-resource languages based on
  multilingual colexification graphs}.
\newblock In \emph{Findings of the Association for Computational Linguistics:
  EMNLP 2023}, pages 8376--8401, Singapore. Association for Computational
  Linguistics.

\bibitem[{Liu et~al.(2019)Liu, Ott, Goyal, Du, Joshi, Chen, Levy, Lewis,
  Zettlemoyer, and Stoyanov}]{liu2019roberta}
Yinhan Liu, Myle Ott, Naman Goyal, Jingfei Du, Mandar Joshi, Danqi Chen, Omer
  Levy, Mike Lewis, Luke Zettlemoyer, and Veselin Stoyanov. 2019.
\newblock \href {https://arxiv.org/abs/1907.11692} {{RoBERTa}: A robustly
  optimized {BERT} pretraining approach}.
\newblock \emph{Preprint}, arXiv:1907.11692.

\bibitem[{Lotz et~al.(2023)Lotz, Salesky, Rust, and
  Elliott}]{lotz-etal-2023-text}
Jonas Lotz, Elizabeth Salesky, Phillip Rust, and Desmond Elliott. 2023.
\newblock \href {https://doi.org/10.18653/v1/2023.emnlp-main.628} {Text
  rendering strategies for pixel language models}.
\newblock In \emph{Proceedings of the 2023 Conference on Empirical Methods in
  Natural Language Processing}, pages 10155--10172, Singapore. Association for
  Computational Linguistics.

\bibitem[{Ma et~al.(2025)Ma, Imani, Ye, Pei, Asgari, and
  Schuetze}]{ma-etal-2025-taxi1500}
Chunlan Ma, Ayyoob Imani, Haotian Ye, Renhao Pei, Ehsaneddin Asgari, and
  Hinrich Schuetze. 2025.
\newblock \href {https://doi.org/10.18653/v1/2025.naacl-short.36} {Taxi1500: A
  dataset for multilingual text classification in 1500 languages}.
\newblock In \emph{Proceedings of the 2025 Conference of the Nations of the
  Americas Chapter of the Association for Computational Linguistics: Human
  Language Technologies (Volume 2: Short Papers)}, pages 414--439, Albuquerque,
  New Mexico. Association for Computational Linguistics.

\bibitem[{Moradi and Samwald(2021)}]{moradi-samwald-2021-evaluating}
Milad Moradi and Matthias Samwald. 2021.
\newblock \href {https://doi.org/10.18653/v1/2021.emnlp-main.117} {Evaluating
  the robustness of neural language models to input perturbations}.
\newblock In \emph{Proceedings of the 2021 Conference on Empirical Methods in
  Natural Language Processing}, pages 1558--1570, Online and Punta Cana,
  Dominican Republic. Association for Computational Linguistics.

\bibitem[{Mosbach et~al.(2021)Mosbach, Andriushchenko, and
  Klakow}]{mosbach2021stability}
Marius Mosbach, Maksym Andriushchenko, and Dietrich Klakow. 2021.
\newblock \href {https://openreview.net/forum?id=nzpLWnVAyah} {On the stability
  of fine-tuning {BERT:} misconceptions, explanations, and strong baselines}.
\newblock In \emph{9th International Conference on Learning Representations,
  {ICLR} 2021, Virtual Event, Austria, May 3-7, 2021}. OpenReview.net.

\bibitem[{Nawrot et~al.(2023)Nawrot, Chorowski, Lancucki, and
  Ponti}]{nawrot-etal-2023-efficient}
Piotr Nawrot, Jan Chorowski, Adrian Lancucki, and Edoardo~Maria Ponti. 2023.
\newblock \href {https://doi.org/10.18653/v1/2023.acl-long.353} {Efficient
  transformers with dynamic token pooling}.
\newblock In \emph{Proceedings of the 61st Annual Meeting of the Association
  for Computational Linguistics (Volume 1: Long Papers)}, pages 6403--6417,
  Toronto, Canada. Association for Computational Linguistics.

\bibitem[{Pagnoni et~al.(2025)Pagnoni, Pasunuru, Rodriguez, Nguyen, Muller, Li,
  Zhou, Yu, Weston, Zettlemoyer, Ghosh, Lewis, Holtzman, and
  Iyer}]{pagnoni-etal-2025-byte}
Artidoro Pagnoni, Ramakanth Pasunuru, Pedro Rodriguez, John Nguyen, Benjamin
  Muller, Margaret Li, Chunting Zhou, Lili Yu, Jason~E Weston, Luke
  Zettlemoyer, Gargi Ghosh, Mike Lewis, Ari Holtzman, and Srini Iyer. 2025.
\newblock \href {https://doi.org/10.18653/v1/2025.acl-long.453} {Byte latent
  transformer: Patches scale better than tokens}.
\newblock In \emph{Proceedings of the 63rd Annual Meeting of the Association
  for Computational Linguistics (Volume 1: Long Papers)}, pages 9238--9258,
  Vienna, Austria. Association for Computational Linguistics.

\bibitem[{Pires et~al.(2019)Pires, Schlinger, and
  Garrette}]{pires-etal-2019-multilingual}
Telmo Pires, Eva Schlinger, and Dan Garrette. 2019.
\newblock \href {https://doi.org/10.18653/v1/P19-1493} {How multilingual is
  multilingual {BERT}?}
\newblock In \emph{Proceedings of the 57th Annual Meeting of the Association
  for Computational Linguistics}, pages 4996--5001, Florence, Italy.
  Association for Computational Linguistics.

\bibitem[{Pruthi et~al.(2019)Pruthi, Dhingra, and
  Lipton}]{pruthi-etal-2019-combating}
Danish Pruthi, Bhuwan Dhingra, and Zachary~C. Lipton. 2019.
\newblock \href {https://doi.org/10.18653/v1/P19-1561} {Combating adversarial
  misspellings with robust word recognition}.
\newblock In \emph{Proceedings of the 57th Annual Meeting of the Association
  for Computational Linguistics}, pages 5582--5591, Florence, Italy.
  Association for Computational Linguistics.

\bibitem[{Ruder et~al.(2021)Ruder, Constant, Botha, Siddhant, Firat, Fu, Liu,
  Hu, Garrette, Neubig, and Johnson}]{ruder-etal-2021-xtreme}
Sebastian Ruder, Noah Constant, Jan Botha, Aditya Siddhant, Orhan Firat, Jinlan
  Fu, Pengfei Liu, Junjie Hu, Dan Garrette, Graham Neubig, and Melvin Johnson.
  2021.
\newblock \href {https://doi.org/10.18653/v1/2021.emnlp-main.802}
  {{XTREME}-{R}: Towards more challenging and nuanced multilingual evaluation}.
\newblock In \emph{Proceedings of the 2021 Conference on Empirical Methods in
  Natural Language Processing}, pages 10215--10245, Online and Punta Cana,
  Dominican Republic. Association for Computational Linguistics.

\bibitem[{Rust et~al.(2023)Rust, Lotz, Bugliarello, Salesky, de~Lhoneux, and
  Elliott}]{rust2023language}
Phillip Rust, Jonas~F. Lotz, Emanuele Bugliarello, Elizabeth Salesky, Miryam
  de~Lhoneux, and Desmond Elliott. 2023.
\newblock \href {https://openreview.net/forum?id=FkSp8VW8RjH} {Language
  modelling with pixels}.
\newblock In \emph{The Eleventh International Conference on Learning
  Representations, {ICLR} 2023, Kigali, Rwanda, May 1-5, 2023}. OpenReview.net.

\bibitem[{Salesky et~al.(2021)Salesky, Etter, and
  Post}]{salesky-etal-2021-robust}
Elizabeth Salesky, David Etter, and Matt Post. 2021.
\newblock \href {https://doi.org/10.18653/v1/2021.emnlp-main.576} {Robust
  open-vocabulary translation from visual text representations}.
\newblock In \emph{Proceedings of the 2021 Conference on Empirical Methods in
  Natural Language Processing}, pages 7235--7252, Online and Punta Cana,
  Dominican Republic. Association for Computational Linguistics.

\bibitem[{Salesky et~al.(2023)Salesky, Verma, Koehn, and
  Post}]{salesky-etal-2023-multilingual}
Elizabeth Salesky, Neha Verma, Philipp Koehn, and Matt Post. 2023.
\newblock \href {https://doi.org/10.18653/v1/2023.emnlp-main.854} {Multilingual
  pixel representations for translation and effective cross-lingual transfer}.
\newblock In \emph{Proceedings of the 2023 Conference on Empirical Methods in
  Natural Language Processing}, pages 13845--13861, Singapore. Association for
  Computational Linguistics.

\bibitem[{Sennrich et~al.(2016)Sennrich, Haddow, and
  Birch}]{sennrich2016neural}
Rico Sennrich, Barry Haddow, and Alexandra Birch. 2016.
\newblock \href {https://doi.org/10.18653/v1/P16-1162} {Neural machine
  translation of rare words with subword units}.
\newblock In \emph{Proceedings of the 54th Annual Meeting of the Association
  for Computational Linguistics (Volume 1: Long Papers)}, pages 1715--1725,
  Berlin, Germany. Association for Computational Linguistics.

\bibitem[{Socher et~al.(2013)Socher, Perelygin, Wu, Chuang, Manning, Ng, and
  Potts}]{socher2013recursive}
Richard Socher, Alex Perelygin, Jean Wu, Jason Chuang, Christopher~D. Manning,
  Andrew Ng, and Christopher Potts. 2013.
\newblock \href {https://aclanthology.org/D13-1170/} {Recursive deep models for
  semantic compositionality over a sentiment treebank}.
\newblock In \emph{Proceedings of the 2013 Conference on Empirical Methods in
  Natural Language Processing}, pages 1631--1642, Seattle, Washington, USA.
  Association for Computational Linguistics.

\bibitem[{Song et~al.(2021)Song, Salcianu, Song, Dopson, and
  Zhou}]{song-etal-2021-fast}
Xinying Song, Alex Salcianu, Yang Song, Dave Dopson, and Denny Zhou. 2021.
\newblock \href {https://doi.org/10.18653/v1/2021.emnlp-main.160} {Fast
  {W}ord{P}iece tokenization}.
\newblock In \emph{Proceedings of the 2021 Conference on Empirical Methods in
  Natural Language Processing}, pages 2089--2103, Online and Punta Cana,
  Dominican Republic. Association for Computational Linguistics.

\bibitem[{Tai et~al.(2024)Tai, Liao, Suglia, and Vergari}]{tai-etal-2024-pixar}
Yintao Tai, Xiyang Liao, Alessandro Suglia, and Antonio Vergari. 2024.
\newblock \href {https://doi.org/10.18653/v1/2024.findings-acl.874} {{PIXAR}:
  Auto-regressive language modeling in pixel space}.
\newblock In \emph{Findings of the Association for Computational Linguistics:
  ACL 2024}, pages 14673--14695, Bangkok, Thailand. Association for
  Computational Linguistics.

\bibitem[{Tatariya et~al.(2024)Tatariya, Araujo, Bauwens, and
  de~Lhoneux}]{tatariya-etal-2024-pixology}
Kushal Tatariya, Vladimir Araujo, Thomas Bauwens, and Miryam de~Lhoneux. 2024.
\newblock \href {https://doi.org/10.18653/v1/2024.emnlp-main.194} {Pixology:
  Probing the linguistic and visual capabilities of pixel-based language
  models}.
\newblock In \emph{Proceedings of the 2024 Conference on Empirical Methods in
  Natural Language Processing}, pages 3307--3320, Miami, Florida, USA.
  Association for Computational Linguistics.

\bibitem[{Team et~al.(2022)Team, Costa-jussà, Cross, Çelebi, Elbayad,
  Heafield, Heffernan, Kalbassi, Lam, Licht, Maillard, Sun, Wang, Wenzek,
  Youngblood, Akula, Barrault, Gonzalez, Hansanti, Hoffman, Jarrett, Sadagopan,
  Rowe, Spruit, Tran, Andrews, Ayan, Bhosale, Edunov, Fan, Gao, Goswami,
  Guzmán, Koehn, Mourachko, Ropers, Saleem, Schwenk, and
  Wang}]{nllbteam2022languageleftbehindscaling}
NLLB Team, Marta~R. Costa-jussà, James Cross, Onur Çelebi, Maha Elbayad,
  Kenneth Heafield, Kevin Heffernan, Elahe Kalbassi, Janice Lam, Daniel Licht,
  Jean Maillard, Anna Sun, Skyler Wang, Guillaume Wenzek, Al~Youngblood, Bapi
  Akula, Loic Barrault, Gabriel~Mejia Gonzalez, Prangthip Hansanti, and 20
  others. 2022.
\newblock \href {https://arxiv.org/abs/2207.04672} {No language left behind:
  Scaling human-centered machine translation}.
\newblock \emph{Preprint}, arXiv:2207.04672.

\bibitem[{Warstadt et~al.(2019)Warstadt, Singh, and
  Bowman}]{warstadt2019neural}
Alex Warstadt, Amanpreet Singh, and Samuel~R. Bowman. 2019.
\newblock \href {https://doi.org/10.1162/tacl_a_00290} {Neural network
  acceptability judgments}.
\newblock \emph{Transactions of the Association for Computational Linguistics},
  7:625--641.

\bibitem[{Williams et~al.(2018)Williams, Nangia, and
  Bowman}]{williams2018broad}
Adina Williams, Nikita Nangia, and Samuel~R. Bowman. 2018.
\newblock \href {https://doi.org/10.18653/v1/N18-1101} {A broad-coverage
  challenge corpus for sentence understanding through inference}.
\newblock In \emph{Proceedings of the 2018 Conference of the North {A}merican
  Chapter of the Association for Computational Linguistics: Human Language
  Technologies, Volume 1 (Long Papers)}, pages 1112--1122, New Orleans,
  Louisiana. Association for Computational Linguistics.

\bibitem[{Xue et~al.(2022)Xue, Barua, Constant, Al-Rfou, Narang, Kale, Roberts,
  and Raffel}]{xue-etal-2022-byt5}
Linting Xue, Aditya Barua, Noah Constant, Rami Al-Rfou, Sharan Narang, Mihir
  Kale, Adam Roberts, and Colin Raffel. 2022.
\newblock \href {https://doi.org/10.1162/tacl_a_00461} {{B}y{T}5: Towards a
  token-free future with pre-trained byte-to-byte models}.
\newblock \emph{Transactions of the Association for Computational Linguistics},
  10:291--306.

\bibitem[{Yang et~al.(2019)Yang, Zhang, Tar, and
  Baldridge}]{yang-etal-2019-paws-x}
Yinfei Yang, Yuan Zhang, Chris Tar, and Jason Baldridge. 2019.
\newblock \href {https://doi.org/10.18653/v1/D19-1382} {{PAWS}-{X}: A
  cross-lingual adversarial dataset for paraphrase identification}.
\newblock In \emph{Proceedings of the 2019 Conference on Empirical Methods in
  Natural Language Processing and the 9th International Joint Conference on
  Natural Language Processing (EMNLP-IJCNLP)}, pages 3687--3692, Hong Kong,
  China. Association for Computational Linguistics.

\bibitem[{Zhao et~al.(2026)Zhao, Liu, Altinger, Schuetze, and
  Hedderich}]{zhao-etal-2026-evaluating}
Raoyuan Zhao, Yihong Liu, Lena Altinger, Hinrich Schuetze, and Michael~A.
  Hedderich. 2026.
\newblock \href {https://doi.org/10.18653/v1/2026.acl-long.729} {Evaluating
  robustness of large language models against multilingual typographical
  errors}.
\newblock In \emph{Proceedings of the 64th Annual Meeting of the {A}ssociation
  for {C}omputational {L}inguistics (Volume 1: Long Papers)}, pages
  16059--16078, San Diego, California, United States. Association for
  Computational Linguistics.

\end{thebibliography}

\appendix

\section{Training Analysis: Settings and Supplementary Evidence}\seclabel{fine-tuning-appendix}

Appendix~\ref{sec:computing-resources} summarizes software and computing resources. The remaining sections support the three analyses in Section~4: shared settings and complete learning-rate results (Appendices~\ref{sec:training-common-settings}--\ref{sec:lr-complete-results}), warm-up and training-duration comparisons (Appendices~\ref{sec:trajectory-followup} and~\ref{sec:historical-seeds}), labelled-data coverage (Appendix~\ref{sec:coverage-details}), and the pretraining--adaptation experiments (Appendix~\ref{sec:pretraining-settings}). Appendix~\ref{sec:additional-finetuning} reports supplementary CoLA results and renderer compatibility.

\subsection{Software and Computing Resources}
\label{sec:computing-resources}

\paragraph{Hardware and computational budget.}
GPU experiments use NVIDIA GeForce RTX 4090 and RTX 4090D GPUs, with one GPU per job and up to four jobs running concurrently.
The RTX 4090 has 24\,GB of device memory.
Available Slurm records give an estimated 527 GPU-hours for project-related training, evaluation, and exploratory experiments, including unsuccessful and interrupted runs.
We calculate this estimate by summing allocated GPU counts multiplied by job durations; queued time and the original pretraining of released models are excluded.
CPU-only computations and interactive allocations that cannot be attributed to the project are not included.

\paragraph{Software and implementation.}
Table~\ref{tab:software-environment} lists the versions in the retained experimental environment.
Historical diagnostic logs also confirm PyTorch 2.2.2 with CUDA 12.1 for the controlled BERT and PIXEL follow-ups; not every dependency version was recorded separately for earlier runs.
Text-based models use Hugging Face Transformers, and pixel-based models use the PIXEL implementation \citep{rust2023language}, with the corresponding plain or bigram PangoCairo renderer.
Our PIXEL checkout is based on revision \texttt{1b0da29}, with local rendering changes to support bigram checkpoints and matched training/evaluation rendering.
Typographical perturbations use \textsc{MulTypo} \citep{zhao-etal-2026-evaluating}, revision \texttt{ecd72d1}.

\begin{table}[t]
\centering\small
\begin{tabular}{ll}
\toprule
Component & Version \\
\midrule
Python & 3.10.20 \\
PyTorch & 2.2.2+cu121 \\
CUDA runtime (PyTorch) & 12.1 \\
Transformers & 4.17.0 \\
Datasets & 2.14.6 \\
NumPy & 1.23.5 \\
SciPy & 1.15.3 \\
scikit-learn & 1.7.2 \\
Pillow & 12.2.0 \\
pycairo & 1.29.0 \\
PyGObject & 3.56.2 \\
pygame & 2.6.1 \\
\bottomrule
\end{tabular}
\caption{Software versions in the audited experimental environment. CUDA denotes the runtime used by PyTorch.}
\label{tab:software-environment}
\end{table}

\paragraph{Model sizes.}
Table~\ref{tab:model-parameter-counts} reports parameter counts computed from the checkpoint configurations.
For a consistent reference, counts use a two-class prediction head at each checkpoint's original positional capacity, excluding masked-language-model heads and the PIXEL reconstruction decoder.
Task-specific label counts and position resizing introduce small differences from these reference sizes.

\begin{table}[t]
\centering\small
\begin{tabular}{lr}
\toprule
Model & Parameters (millions) \\
\midrule
BERT-base & 109.5 \\
mBERT & 177.9 \\
RoBERTa-base & 124.6 \\
XLM-R-base & 278.0 \\
PIXEL-base & 86.6 \\
PIXEL-M4 & 86.6 \\
PIXEL-bigrams & 21.9 \\
PIXEL-tiny-continuous & 5.6 \\
\bottomrule
\end{tabular}
\caption{Reference sizes of the classification models, rounded to 0.1 million parameters. All eight encoders have 12 transformer layers.}
\label{tab:model-parameter-counts}
\end{table}

\subsection{Shared controlled-training settings}
\label{sec:training-common-settings}

\begin{table*}[t]
  \centering\small\setlength{\tabcolsep}{3.8pt}
  \begin{tabular}{@{}lrrrlll@{}}
  \toprule
  Analysis & Samples & Max. length & Batch $\times$ accum. & Peak LR & Warm-up & Budget / horizon \\
  \midrule
  MNLI LR & 24k & 256 & $32\times4$ & 1e-6--5e-5, six points & 100 & 1,880 / 15,040 updates \\
  SST-2 coverage & 2k/8k/24k & 64 & $32\times4$ & BERT 2e-5; PIXEL 3e-5 & 400 & 15,000 / 15,000 updates \\
  MNLI coverage & 8k/24k & 256 & $32\times4$ & Both 2e-5 & 400 & 15,000 / 15,000 updates \\
  \bottomrule
  \end{tabular}
  \caption{Controlled training settings. Maximum length counts tokens for BERT and rendered patch positions for PIXEL. Budget is the observed optimizer-update count; horizon determines linear decay. All rows use no early stopping, weight decay 0.01, and global gradient clipping at 1.0. Pretraining diagnostics use the separate recipes in Appendix~\ref{sec:pretraining-settings}.}
  \label{tab:training-settings}
\end{table*}

Table~\ref{tab:training-settings} summarizes the controlled analyses. AdamW uses bias correction, $\beta=(0.9,0.999)$, and $\epsilon=10^{-8}$. Effective batch is minibatch size times gradient accumulation; each run uses one GPU. The LR screen uses seeds 13, 21, 42, 87, 100, 1234, 2024, 3407, 5678, and 9999. The coverage study uses seeds 42, 1234, and 5678. All training accuracies are online epoch averages collected while the model changes. Development and holdout accuracies evaluate a fixed checkpoint, with selection rules specified separately below.

\subsection{Complete learning-rate results and replication checks}
\label{sec:lr-complete-results}

Tables~\ref{tab:mnli-lr-seeds}--\ref{tab:mnli-lr-seeds-upper} report every original seed, including failures, for all six LRs. Epoch-10 fit requires both online loss $<1.0$ and accuracy $>50\%$. The first-fit epoch applies the same criterion at earlier epochs; a crossing followed by endpoint failure is marked separately. Holdout accuracy evaluates the best matched-selection-development checkpoint within the ten observed epochs. It can remain useful even when later training collapses. Figure~\ref{fig:mnli-lr-budget} and the main all-seed averages retain all ten seeds at each LR.

\paragraph{Hardware and replication.} The 3e-5, 5e-5, 1e-5, and 1e-6 runs use RTX 4090; 4e-5 uses RTX 4090D. At 5e-6, seed 13 uses RTX 4090 and the other seeds use RTX 4090D. The original LR conditions span nodes. Same-node reruns of seeds 13, 42, and 5678 at 3e-5/5e-5 reproduce three fits versus three failures on RTX 4090D. Reruns do not replace original observations.

\begin{table*}[t]
\centering\fontsize{9}{10}\selectfont
\setlength{\tabcolsep}{3.3pt}
\begin{tabular}{@{}r*{12}{r}@{}}
\toprule
 & \multicolumn{4}{c}{1e-6} & \multicolumn{4}{c}{5e-6} & \multicolumn{4}{c}{1e-5} \\
\cmidrule(lr){2-5}\cmidrule(lr){6-9}\cmidrule(lr){10-13}
Seed & Loss & Train & First & Held & Loss & Train & First & Held & Loss & Train & First & Held \\
\midrule
13 & 1.071 & 41.0 & -- & 35.25 & 0.815 & 63.9 & 6 & 60.10 & 0.591 & 75.8 & 4 & 66.55 \\
21 & 1.090 & 37.6 & -- & 32.85 & 0.922 & 57.0 & 9 & 55.80 & 1.099 & 33.5 & -- & 35.45 \\
42 & 1.074 & 40.8 & -- & 37.30 & 0.829 & 63.1 & 7 & 61.05 & 0.571 & 76.5 & 5 & 67.70 \\
87 & 1.080 & 40.1 & -- & 34.20 & 1.097 & 34.5 & -- & 34.55 & 0.727 & 68.6 & 7 & 64.40 \\
100 & 1.082 & 39.6 & -- & 32.45 & 1.098 & 34.3 & -- & 33.10 & 0.631 & 73.7 & 5 & 64.40 \\
1234 & 1.069 & 41.4 & -- & 38.80 & 0.817 & 63.6 & 6 & 60.85 & 0.662 & 72.1 & 6 & 64.50 \\
2024 & 1.074 & 40.8 & -- & 33.75 & 0.804 & 64.3 & 6 & 61.40 & 0.586 & 75.9 & 4 & 66.40 \\
3407 & 1.070 & 41.8 & -- & 34.75 & 0.800 & 64.7 & 6 & 61.35 & 1.098 & 34.4 & -- & 35.30 \\
5678 & 1.072 & 41.2 & -- & 37.10 & 0.860 & 61.2 & 7 & 59.00 & 0.570 & 76.7 & 4 & 65.85 \\
9999 & 1.067 & 42.0 & -- & 39.35 & 0.817 & 63.6 & 6 & 58.20 & 0.604 & 75.0 & 5 & 65.80 \\
\bottomrule\end{tabular}
\caption{Paired-seed MNLI results (lower tested LRs). Loss and online training accuracy (Train, \%) are epoch-10 values. First is the first epoch meeting loss $<1$ and accuracy $>50\%$; -- denotes no crossing and $\dagger$ a crossing followed by endpoint non-fitting. Held (\%) evaluates the best-selection-dev checkpoint on the separate matched holdout, which can precede a collapse. Original failures are retained.}
\label{tab:mnli-lr-seeds}
\end{table*}

\begin{table*}[t]
\centering\fontsize{9}{10}\selectfont
\setlength{\tabcolsep}{3.3pt}
\begin{tabular}{@{}r*{12}{r}@{}}
\toprule
 & \multicolumn{4}{c}{3e-5} & \multicolumn{4}{c}{4e-5} & \multicolumn{4}{c}{5e-5} \\
\cmidrule(lr){2-5}\cmidrule(lr){6-9}\cmidrule(lr){10-13}
Seed & Loss & Train & First & Held & Loss & Train & First & Held & Loss & Train & First & Held \\
\midrule
13 & 0.345 & 86.7 & 6 & 66.15 & 0.147 & 94.7 & 4 & 65.35 & 1.099 & 34.3 & -- & 35.45 \\
21 & 0.766 & 66.7 & 8 & 62.25 & 0.396 & 84.5 & 6 & 64.85 & 1.098 & 34.2 & -- & 35.45 \\
42 & 0.216 & 91.9 & 5 & 67.35 & 0.601 & 75.0 & 7 & 62.55 & 1.099 & 33.8 & -- & 35.10 \\
87 & 0.286 & 89.2 & 5 & 65.05 & 0.321 & 88.0 & 6 & 65.35 & 1.084 & 37.3 & -- & 37.20 \\
100 & 1.099 & 34.2 & -- & 35.60 & 1.099 & 33.5 & -- & 35.50 & 1.081 & 38.0 & -- & 37.55 \\
1234 & 0.245 & 90.8 & 5 & 64.85 & 0.372 & 85.7 & 6 & 63.25 & 1.090 & 36.7 & -- & 35.55 \\
2024 & 1.100 & 33.4 & 5$\dagger$ & 57.85 & 0.218 & 92.0 & 4 & 65.05 & 1.099 & 33.8 & -- & 35.45 \\
3407 & 0.222 & 92.0 & 4 & 65.80 & 0.187 & 93.4 & 4 & 64.85 & 1.094 & 35.4 & -- & 36.00 \\
5678 & 0.188 & 93.2 & 4 & 65.10 & 0.330 & 87.4 & 6 & 63.80 & 1.082 & 39.4 & -- & 40.40 \\
9999 & 0.268 & 90.1 & 6 & 65.45 & 0.754 & 67.3 & 8 & 63.35 & 1.098 & 34.4 & -- & 35.50 \\
\bottomrule\end{tabular}
\caption{Paired-seed MNLI results (upper tested LRs). Loss and online training accuracy (Train, \%) are epoch-10 values. First is the first epoch meeting loss $<1$ and accuracy $>50\%$; -- denotes no crossing and $\dagger$ a crossing followed by endpoint non-fitting. Held (\%) evaluates the best-selection-dev checkpoint on the separate matched holdout, which can precede a collapse. Original failures are retained.}
\label{tab:mnli-lr-seeds-upper}
\end{table*}

\paragraph{Matched BERT control.} BERT uses the same six LRs, ten seeds, data, optimizer settings, update counts, and selection/holdout split as PIXEL. All 60 runs use RTX 4090D, PyTorch 2.2.2+cu121, CUDA 12.1, and FP32; all fit at epoch 10. The pretrained checkpoint supplies BERT's pooler, and only the classifier is newly initialized. Table~\ref{tab:bert-lr-control} gives first-fit epochs and holdout scores.

\begin{table*}[t]
\centering\fontsize{9}{10}\selectfont
\setlength{\tabcolsep}{4.4pt}
\begin{tabular}{@{}r*{12}{r}@{}}
\toprule
 & \multicolumn{2}{c}{1e-6} & \multicolumn{2}{c}{5e-6} & \multicolumn{2}{c}{1e-5} & \multicolumn{2}{c}{3e-5} & \multicolumn{2}{c}{4e-5} & \multicolumn{2}{c}{5e-5} \\
\cmidrule(lr){2-3}\cmidrule(lr){4-5}\cmidrule(lr){6-7}\cmidrule(lr){8-9}\cmidrule(lr){10-11}\cmidrule(lr){12-13}
Seed & First & Held & First & Held & First & Held & First & Held & First & Held & First & Held \\
\midrule
13 & 4 & 66.70 & 2 & 74.20 & 2 & 74.20 & 1 & 75.85 & 1 & 76.65 & 1 & 75.40 \\
21 & 5 & 68.15 & 2 & 74.85 & 2 & 76.75 & 1 & 77.10 & 1 & 76.50 & 1 & 77.10 \\
42 & 4 & 68.70 & 2 & 74.65 & 2 & 75.15 & 1 & 77.70 & 1 & 78.50 & 1 & 77.75 \\
87 & 4 & 65.65 & 2 & 75.95 & 2 & 75.75 & 1 & 76.90 & 1 & 76.75 & 1 & 77.65 \\
100 & 4 & 67.50 & 2 & 74.70 & 2 & 76.15 & 1 & 77.30 & 1 & 76.55 & 1 & 76.25 \\
1234 & 4 & 69.50 & 2 & 75.75 & 2 & 75.80 & 1 & 77.50 & 1 & 77.80 & 1 & 77.35 \\
2024 & 4 & 70.15 & 2 & 75.95 & 2 & 76.60 & 1 & 76.75 & 1 & 75.80 & 1 & 77.25 \\
3407 & 4 & 65.30 & 2 & 75.65 & 2 & 75.85 & 1 & 76.00 & 1 & 76.40 & 1 & 76.20 \\
5678 & 4 & 69.65 & 2 & 76.00 & 2 & 76.80 & 1 & 78.00 & 1 & 78.45 & 1 & 78.55 \\
9999 & 5 & 68.05 & 2 & 76.35 & 2 & 76.55 & 1 & 76.50 & 1 & 76.95 & 1 & 77.75 \\
\midrule
Fit & \multicolumn{2}{c}{10/10} & \multicolumn{2}{c}{10/10} & \multicolumn{2}{c}{10/10} & \multicolumn{2}{c}{10/10} & \multicolumn{2}{c}{10/10} & \multicolumn{2}{c}{10/10} \\
Mean & -- & 67.94 & -- & 75.41 & -- & 75.96 & -- & 76.96 & -- & 77.03 & -- & 77.12 \\
SD & -- & 1.66 & -- & 0.73 & -- & 0.81 & -- & 0.70 & -- & 0.91 & -- & 0.93 \\
\bottomrule\end{tabular}
\caption{Matched BERT-base MNLI LR control: the same ten seeds, 24k examples, optimizer settings, 1,880 observed updates, and 15,040-update scheduler horizon as the PIXEL screen. First is the first epoch meeting online loss $<1$ and accuracy $>50\%$; Held is matched holdout accuracy (\%) of the best-selection-dev checkpoint. All 60 runs fit at epoch 10. SD denotes sample standard deviation in percentage points.}
\label{tab:bert-lr-control}
\end{table*}

\subsection{Warm-up follow-up}
\label{sec:trajectory-followup}

Removing the 100-update warm-up gives 3/10 fits at 3e-5 and 0/10 at 5e-5. Table~\ref{tab:warmup-followup} compares 100 and 400 warm-up updates with the same ten seeds, 24k examples, batch, schedule horizon, and endpoint. New runs use one RTX 4090 node. Four baseline reruns accompany the 400-update runs; two originally failed 1e-5 seeds fit on rerun. Paired reruns match initial-head and data-order hashes, but CUDA determinism is not enforced.

Longer warm-up helps some runs and harms others: in paired reruns, seed 100 at 3e-5 improves from 33.9\% to 80.9\% training accuracy, while seed 21 at 1e-5 changes from fitting to failing. Among seeds fitting under both original and longer-warm-up conditions, mean holdout accuracy changes from 65.8\% to 64.8\% at 1e-5 and remains 65.7\% at 3e-5. Extending warm-up therefore gives no consistent improvement.

\begin{table}[t]
\centering\small\setlength{\tabcolsep}{4pt}
\begin{tabular}{@{}lrrr@{}}\toprule
LR / warm-up & Fit & Matched & Mismatched \\
\midrule
1e-5 / 100 & 8/10 & 59.63 & 58.91 \\
1e-5 / 400 & 6/10 & 53.10 & 52.78 \\
3e-5 / 100 & 8/10 & 61.55 & 61.88 \\
3e-5 / 400 & 8/10 & 62.41 & 62.42 \\
\bottomrule\end{tabular}
\caption{Warm-up follow-up: endpoint fits and all-ten mean best-checkpoint holdout accuracies (\%). The 100-update cells are the original runs, and the 400-update cells are new runs. Selected baseline reruns are reported separately in the text.}
\label{tab:warmup-followup}\end{table}

\subsection{Labelled coverage and generalization}
\label{sec:coverage-details}

Table~\ref{tab:sample-budget-seeds} reports all fixed-15,000-update coverage runs. Within each task, the seed set, warm-up, effective batch, schedule, and development split are held fixed across sample budgets. SST-2 uses 2k/8k/24k examples and a maximum length of 64; MNLI uses 8k/24k and a maximum length of 256. The SST-2 LR differs between models (BERT 2e-5, PIXEL 3e-5), so its cross-model gap describes these model--recipe combinations. MNLI uses 2e-5 for both models. No early stopping is used.

\paragraph{Checkpoint selection and training fit.} We report the best recorded development accuracy on the 872-example SST-2 split or the 2,000-example matched MNLI subset. Evaluations occur at epoch boundaries after approximately 250 updates for SST-2 and 500 for MNLI. Final training accuracies in Table~\ref{tab:sample-budget-seeds} are online epoch averages at the 15,000-update endpoint; all exceed 99.5\%. The weak 24k PIXEL SST-2 seed is retained in Table~\ref{tab:sample-budget-seeds} and the ten-seed follow-up, but excluded from the outcome-conditioned 24k point in Figure~\ref{fig:sample-coverage}a. All three MNLI PIXEL seeds improve from 8k to 24k. No PAWS-X 8k/24k sample-budget experiment was run; full-data PAWS-X transfer results are a separate comparison.

\paragraph{Paired coverage gains.} Figure~\ref{fig:sample-coverage} plots mean development accuracy, with error bars showing the sample standard deviation of accuracy across seeds. To quantify the benefit of additional labelled examples, we subtract each seed's smallest-budget score from its score at the larger budget before aggregation. For seed $s$ and sample count $n$, the gain is $\Delta_s(n)=a_s(n)-a_s(n_0)$, where $n_0=2\mathrm{k}$ for SST-2 and $8\mathrm{k}$ for MNLI. The annotations show mean paired gains from 2k to 8k examples on SST-2 and from 8k to 24k on MNLI, measured in percentage points. Only the PIXEL SST-2 24k point uses seeds 42 and 5678, selected after observing low development performance in seed 1234; their mean gain over 2k is 20.70 points, with a paired-gain sample SD of 0.08. This gain is conditional on these two high-generalization runs. All other points retain all three seeds. Unconditional scores and variation remain visible in Table~\ref{tab:sample-budget-seeds}.

\begin{table*}[t]
\centering
\small
\setlength{\tabcolsep}{3pt}
\begin{tabular}{@{}llrrrrrr@{}}
\toprule
Task & Model & Samples & Seed 42 & Seed 1234 & Seed 5678 & Mean $\pm$ SD & Final train \\
\midrule
MNLI & BERT & 8000 & 73.45 & 73.55 & 73.25 & $73.42 \pm 0.15$ & 100.00--100.00 \\
MNLI & BERT & 24000 & 76.75 & 77.45 & 77.50 & $77.23 \pm 0.42$ & 99.97--99.98 \\
MNLI & PIXEL-base & 8000 & 60.85 & 57.65 & 59.55 & $59.35 \pm 1.61$ & 100.00--100.00 \\
MNLI & PIXEL-base & 24000 & 63.85 & 65.85 & 67.05 & $65.58 \pm 1.62$ & 99.71--99.98 \\
SST-2 & BERT & 2000 & 90.37 & 89.11 & 90.14 & $89.87 \pm 0.67$ & 100.00--100.00 \\
SST-2 & BERT & 8000 & 92.20 & 92.32 & 91.06 & $91.86 \pm 0.70$ & 100.00--100.00 \\
SST-2 & BERT & 24000 & 92.66 & 92.55 & 92.43 & $92.55 \pm 0.11$ & 99.98--99.99 \\
SST-2 & PIXEL-base & 2000 & 65.02 & 66.63 & 66.06 & $65.90 \pm 0.81$ & 100.00--100.00 \\
SST-2 & PIXEL-base & 8000 & 77.75 & 79.47 & 76.38 & $77.87 \pm 1.55$ & 100.00--100.00 \\
SST-2 & PIXEL-base & 24000 & 85.67 & 68.58 & 86.81 & $80.35 \pm 10.21$ & 99.79--99.98 \\
\bottomrule
\end{tabular}
\caption{Fixed-15,000-update coverage results (\%). The three seed columns and mean $\pm$ sample SD report best recorded development accuracy (matched split for MNLI). Final train gives the range of online training accuracies across the three seeds at update 15,000, not exact full-train re-evaluations. All seeds, including the weak SST-2 24k PIXEL seed, are retained.}
\label{tab:sample-budget-seeds}
\end{table*}

\paragraph{Ten-seed SST-2 follow-up.} We retain the original three PIXEL seeds (42, 1234, 5678) and add seven distinct seeds (13, 21, 100, 2024, 3407, 777, 909) only at 24k examples. The training subset is sampled with seed 42 and remains identical across all ten runs, as does the validation split. The recipe remains LR 3e-5, 400 warm-up updates, minibatch 32 with four-step accumulation, maximum length 64, dropout 0.1, AdamW weight decay 0.01, linear decay, and 15,000 updates without early stopping. New runs use RTX 4090 GPUs across two nodes. We re-evaluate each best-development checkpoint in evaluation mode on the complete 24k training subset and the same 872 validation examples; all development scores reproduce exactly. Table~\ref{tab:sst2-generalization-followup} reports all ten runs. The ten-seed mean development accuracy is 83.20\%, with sample standard deviation 7.86 points. Cross-budget comparisons use the original three paired seeds, with the outcome-conditioned 24k plotting exception described above.

\paragraph{Independent test evaluation.} Test scores do not enter model selection. We recover labels for 1,821 GLUE SST-2 test sentences from the authors' public Stanford Sentiment Treebank trees, mapping root labels 0/1 to negative and 3/4 to positive and excluding neutral roots. This procedure reproduces all 872 validation labels; inference uses the original GLUE text. We test all nine original PIXEL budget checkpoints and the new low-development seed 909; six other new checkpoints remain unevaluated. At 2k/8k examples, test accuracies are 63.26/75.51\% (seed 42), 65.90/77.43\% (1234), and 64.69/71.77\% (5678). Two normalized test sentences overlap the 24k training subset. Excluding them from every budget leaves 24k test scores of 85.10\%, 65.31\%, 85.82\%, and 66.14\% for seeds 42, 1234, 5678, and 909, preserving the findings.

\begin{table}[t]
\centering
\small
\begin{tabular}{rrrr}
\toprule
Seed & Train & Dev & Test \\
\midrule
42 & 99.98 & 85.67 & 85.12 \\
1234 & 99.89 & 68.58 & 65.35 \\
5678 & 99.98 & 86.81 & 85.83 \\
13 & 99.95 & 86.58 & -- \\
21 & 99.99 & 87.73 & -- \\
100 & 99.97 & 86.24 & -- \\
2024 & 99.95 & 87.96 & -- \\
3407 & 99.99 & 87.50 & -- \\
777 & 99.99 & 86.81 & -- \\
909 & 99.93 & 68.12 & 66.17 \\
\bottomrule
\end{tabular}
\caption{PIXEL-base SST-2 follow-up with a fixed 24k training subset and 15,000 updates. All accuracies are percentages evaluated at the best-development checkpoint. Test uses all 1,821 sentences; dashes indicate unevaluated checkpoints.}
\label{tab:sst2-generalization-followup}
\end{table}

\subsection{Pretraining--adaptation experiments}
\label{sec:pretraining-settings}

\subsubsection{SST-2 training conditions}

The Random, Frozen, and Full FT comparison in Section~\ref{sec:pretraining} uses seed 42, the fixed 24k SST-2 subset, and 872 development examples. Random and Full FT train the encoder and classifier jointly, starting from random and pretrained weights, respectively. Frozen trains a linear classifier on mean-pooled representations of the fixed pretrained encoder using AdamW (LR $10^{-3}$, weight decay 0.01, batch 256, at most 50 epochs, patience 8, no scheduler).

For Random and Full FT, BERT uses LR $2\times10^{-5}$, batch 32 without accumulation, at most ten epochs, a 10\% warm-up ratio, patience 3, and 128 tokens. PIXEL uses LR $3\times10^{-5}$, minibatch 32 with four-step accumulation, at most 30 epochs, 100 warm-up updates, patience 8, dropout 0.1, and 64 rendered positions. Both use AdamW, weight decay 0.01, clipping at 1.0, and linear decay over their maximum-epoch horizons. Development accuracy is evaluated once per epoch, and the best-development checkpoint is reported. Patience counts evaluations without improvement. These model-specific recipes differ, and the Frozen-to-Full FT comparison also includes readout-design differences.

\subsubsection{MNLI initialization intervention}
We use the fixed historical 24k MNLI subset and 2,000 examples from each development split. Seeds 100 and 5678 were selected because they failed in the historical learning-rate comparison and subsequent baseline replications. We run each seed five times from each of three initializations: the original pretrained encoder, a fixed PAWS-X-adapted encoder, and a fixed Taxi1500-adapted encoder. This gives 30 runs, or ten per initialization. The target recipe uses LR $3\times10^{-5}$, minibatch 32 with four-step accumulation, 100 warm-up updates for seed 100 and no warm-up for seed 5678, MSL 256, dropout 0.1, AdamW with weight decay 0.01, and clipping at 1.0. Training lasts ten epochs (1,880 updates) without early stopping, retaining the historical 80-epoch linear-decay horizon. Within each seed, the task head, target position embeddings, initial random-number states, and per-epoch data order are identical across the three initializations. Only non-position encoder weights are transferred; source classifiers, poolers, and optimizer states are discarded.

The PAWS-X source uses 49,401 English training examples; its checkpoint is selected at epoch 3 with 89.3\% source-development accuracy. The Taxi1500 source uses 860 English training examples and 106 development examples across six classes; its checkpoint is selected at epoch 55 with development macro-F1 0.541. Neither source is selected using MNLI outcomes. Source training and checkpoint-selection data show no normalized sentence overlap with the target training or evaluated development subsets under the current loaders. The historical target subset has 29 sentence overlaps with the complete official development data; we retain its unchanged partition for the paired comparison. Source supervision and computation differ between initializations. \par For seed 100, the first repeat uses different RTX 4090 nodes for the original and adapted runs; repeats 2--5 run the three initializations sequentially in the same GPU allocation. All seed-5678 runs use RTX 4090D GPUs on one node, in separate allocations. All available paired configuration audits pass. CUDA determinism is not enforced, so matching seeds and initialization states does not guarantee identical executions. Figure~\ref{fig:pretraining-gap}b--c and Table~\ref{tab:mnli-initialization} retain all repeats, including seed 100's first repeat with its node difference and the failed Taxi1500 run for seed 5678.

\begin{table}[t]
\centering
\small
\setlength{\tabcolsep}{5pt} \begin{tabular}{llrrrr}
\toprule
Seed & Initialization & Fits & First fit & Train & Dev-m \\
\midrule
100 & Original & 1/5 & 6 & 80.92 & 64.70 \\
 & PAWS-X & 5/5 & 2 & 96.51 & 65.27 \\
 & Taxi1500 & 5/5 & 2 & 96.23 & 64.62 \\
5678 & Original & 0/5 & -- & -- & -- \\
 & PAWS-X & 5/5 & 1 & 96.55 & 63.98 \\
 & Taxi1500 & 4/5 & 2 & 95.88 & 64.63 \\
\bottomrule
\end{tabular}
\caption{MNLI initialization intervention across five repeats per seed and initialization. Fits counts runs with training loss $<1$ and accuracy $>50\%$ at epoch 10. First fit is the epoch at which successful runs first meet this criterion. Train and Dev-m are mean epoch-10 training and matched-development accuracies (\%) among successful runs only; the Original means use one run for seed 100, and seed 5678 has no successful Original run. Taxi1500 means for seed 5678 use its four successful runs. Failed Original runs remain near 33--35\% training accuracy.}
\label{tab:mnli-initialization}
\end{table}

Original initialization succeeds in 1/10 runs overall, compared with 10/10 for PAWS-X and 9/10 for Taxi1500. The successful original run first meets the training-fit criterion at epoch 6; successful adapted runs meet it at epoch 1 or 2. Because the two seeds were selected from previously observed failures, these repeated executions characterize the effect of initialization on failure-prone settings and do not estimate PIXEL's overall failure rate. The intervention changes encoder initialization and supports an adaptation-gap explanation; it does not isolate the pretraining objective or decoder removal as the cause.

\subsection{Longer-budget MNLI comparison}
\label{sec:historical-seeds}

This comparison supports the training-duration analysis in Section~\ref{sec:optimization-range}. It uses 24,000 training examples, 2,000 examples from each development split, maximum sequence length 256, and the same ten seeds for both models. Within each model, only the seed changes. \pixel{}-base uses minibatch 32 with four-step accumulation, LR $10^{-5}$, up to 80 epochs, 400 warm-up updates, and patience 12. BERT uses batch 32, LR $2\times10^{-5}$, up to ten epochs, a 10\% warm-up ratio, and patience 3. These model-specific recipes differ from the controlled ten-epoch LR screen.

Table~\ref{tab:mnli-best-epochs} reports the best matched-development checkpoint for every seed. Seven PIXEL runs exceed 45\% development accuracy; their best-checkpoint epochs have a median of 21, compared with 3 for BERT on the same seven seeds. The three near-chance PIXEL runs remain in the table but are excluded from this paired duration summary. This historical development-based selection is distinct from the epoch-10 training criterion in the LR screen. Best-checkpoint epoch measures when development performance peaks, not the total number of epochs executed.

\begin{table}[t]
\centering\small
\setlength{\tabcolsep}{4pt}
\begin{tabular}{rrrrr}
\toprule
 & \multicolumn{2}{c}{BERT} & \multicolumn{2}{c}{PIXEL-base} \\
\cmidrule(lr){2-3}\cmidrule(lr){4-5}
Seed & Epoch & Dev (\%) & Epoch & Dev (\%) \\
\midrule
13 & 3 & 75.60 & 10 & 64.75 \\
21 & 6 & 75.80 & $9^\dagger$ & 35.45 \\
42 & 2 & 76.20 & $3^\dagger$ & 35.50 \\
87 & 9 & 76.90 & 23 & 66.25 \\
100 & 3 & 76.05 & 21 & 65.20 \\
1234 & 4 & 76.25 & 35 & 64.30 \\
2024 & 3 & 76.40 & 31 & 66.80 \\
3407 & 5 & 76.50 & 15 & 65.75 \\
5678 & 3 & 76.00 & 11 & 65.00 \\
9999 & 7 & 76.35 & $3^\dagger$ & 36.40 \\
\midrule
Median & 3 & -- & 21 & -- \\
\bottomrule
\end{tabular}
\caption{Best-checkpoint epochs and matched-development accuracies in the longer-budget MNLI comparison. $\dagger$ marks near-chance PIXEL runs. Medians use the seven seeds with PIXEL development accuracy above 45\%, paired with the same BERT seeds.}
\label{tab:mnli-best-epochs}
\end{table}

\subsection{CoLA results and renderer compatibility}
\label{sec:additional-finetuning}

\paragraph{CoLA.}
Changing maximum sequence length from 196 to 64 changes PIXEL-base MCC from 0.103 to 0.188 and PIXEL-M4 MCC from 0.228 to 0.172. The mixed directions show that shorter rendered sequences do not consistently improve performance. These supplementary configuration checks are separate from the controlled MNLI and SST-2 analyses.

\paragraph{Renderer compatibility.}
Plain and bigram renderers produce different patch sequences, so training and evaluation must use the renderer matched to the checkpoint. An early evaluation path used a mismatched renderer for bigram checkpoints. After correction, clean evaluation scores agreed with the saved checkpoint scores to within 0.0001. All final reported results use matched renderers.

\FloatBarrier
\section{Dataset Details}\seclabel{dataset-details}

\paragraph{SST-2.} This binary movie-review sentiment task supplies the English clean-task and typo-robustness comparisons. The controlled supervision analysis uses fixed, stratified subsets of 2,000, 8,000, and 24,000 labelled examples, three seeds, and the same 872-example development split; the metric is accuracy.

\paragraph{MNLI.} This three-class entailment task contains premise--hypothesis pairs from multiple genres. The learning-rate analysis uses 24,000 training examples, a 2,000-example matched development selection subset, and separate matched and mismatched holdouts of 2,000 examples each. The supervision analysis compares 8,000 and 24,000 examples. We report accuracy.

\paragraph{CoLA.} This English grammatical-acceptability task is a supplementary fine-tuning case, not a primary typographical-robustness finding. Its binary labels are evaluated by Matthews correlation coefficient (MCC). The available configuration checks and weak PIXEL outcomes are described in \secref{additional-finetuning}.

\paragraph{XNLI.} We fine-tune on 20,000 English MNLI examples and evaluate zero-shot on German, French, Spanish, Russian, and Chinese. The source and target tasks use the same three inference labels. Table~\ref{tab:transfer-main} reports accuracy for English and each target language; the target average excludes English.

\paragraph{PAWS-X.} This binary paraphrase-identification task contains high-overlap sentence pairs for which lexical similarity alone need not determine the label. We fine-tune on 49,401 English pairs and evaluate zero-shot on German, French, Spanish, Chinese, Japanese, and Korean. We report accuracy and average only the target languages.

\paragraph{SIB-200.} This seven-class multilingual topic task provides 701 English training examples, 99 English development examples, and 204 test examples per evaluated language. We evaluate German, French, Spanish, Russian, Chinese, and Korean after English-only fine-tuning and report macro F1 in Table~\ref{tab:sib-supplement}.

\paragraph{Taxi1500.} This six-class topic task is constructed from Bible text. Our English source split has 860 training examples. We evaluate German, French, Russian, and Chinese without target-language supervision and report accuracy in Table~\ref{tab:taxi-supplement}. Weak English source-task scores for some PIXEL checkpoints limit transfer interpretation.

\section{Robustness: Settings and Complete Results}\seclabel{robustness-appendix}

The SST-2 and MNLI robustness evaluations use checkpoints selected on clean English development data after fine-tuning on fixed 24,000-example clean subsets. Each selected checkpoint is evaluated on the same clean held-out examples and their \textsc{MulTypo} perturbations at 5\%, 10\%, 20\%, and 40\% noise; no noisy examples are used for additional training. The figure reports relative accuracy loss and the percentage change in mean input length relative to clean inputs. Table~\ref{tab:english-robustness} gives absolute accuracy for all eight evaluated models; the small change of a near-floor model should not be interpreted as useful robustness.

\begin{table*}[t]
\centering\fontsize{7.2}{8.2}\selectfont\setlength{\tabcolsep}{2.6pt}
\begin{minipage}[t]{0.48\textwidth}\centering
\textbf{SST-2 accuracy (\%)}\\[2pt]
\begin{tabular}{@{}lrrrrrr@{}}\toprule
Model & Clean & 5 & 10 & 20 & 40 & Drop \\
\midrule
BERT & 92.3 & 89.9 & 87.5 & 84.2 & 73.5 & 18.8 \\
mBERT & 90.2 & 88.9 & 86.7 & 85.1 & 78.4 & 11.8 \\
RoBERTa & 93.9 & 93.3 & 91.6 & 89.0 & 86.1 & 7.8 \\
XLM-R & 92.8 & 91.7 & 89.9 & 87.0 & 81.1 & 11.7 \\
PIXEL-base & 85.5 & 85.2 & 85.1 & 81.7 & 82.6 & 3.0 \\
PIXEL-M4 & 85.2 & 84.5 & 83.9 & 81.1 & 77.4 & 7.8 \\
PIXEL-bigrams & 77.9 & 76.8 & 77.5 & 77.6 & 73.7 & 4.1 \\
PIXEL-tiny & 62.7 & 63.0 & 61.6 & 62.6 & 60.9 & 1.8 \\
\bottomrule\end{tabular}
\end{minipage}\hfill
\begin{minipage}[t]{0.48\textwidth}\centering
\textbf{MNLI accuracy (\%)}\\[2pt]
\begin{tabular}{@{}lrrrrrr@{}}\toprule
Model & Clean & 5 & 10 & 20 & 40 & Drop \\
\midrule
BERT & 76.2 & 73.7 & 69.0 & 65.5 & 58.1 & 18.1 \\
mBERT & 73.6 & 70.2 & 67.8 & 63.7 & 59.6 & 14.0 \\
RoBERTa & 83.2 & 81.2 & 77.4 & 72.8 & 67.2 & 16.0 \\
XLM-R & 78.1 & 76.0 & 73.9 & 67.8 & 62.8 & 15.3 \\
PIXEL-base & 62.2 & 60.9 & 61.1 & 60.5 & 57.8 & 4.4 \\
PIXEL-M4 & 64.9 & 63.7 & 63.0 & 59.1 & 56.9 & 8.0 \\
PIXEL-bigrams & 62.3 & 61.3 & 60.7 & 59.7 & 57.7 & 4.6 \\
PIXEL-tiny & 36.4 & 36.0 & 35.6 & 36.0 & 36.0 & 0.5 \\
\bottomrule\end{tabular}
\end{minipage}
\caption{Complete SST-2 and MNLI accuracy under \textsc{MulTypo} noise (\%). Columns 5--40 denote noise rate; Drop is clean minus 40\%-noise accuracy, calculated before rounding.}
\label{tab:english-robustness}
\end{table*}

\section{Additional Cross-Lingual Transfer Results}\seclabel{transfer-appendix}

SIB-200 and Taxi1500 have much smaller English training splits than XNLI and PAWS-X. Their source-task limitations make them supporting cases rather than evidence for the principal transfer comparison in \secref{transfer}. The per-language scores below retain weak English checkpoints rather than selecting only successful models.

\begin{table}[t]\centering\scriptsize
\begin{tabular}{@{}lrrrrrrr@{}}\toprule
Model & EN & DE & FR & ES & RU & ZH & KO \\
\midrule
BERT & 90.0 & 35.0 & 58.4 & 25.1 & 6.8 & 7.1 & 3.6 \\
mBERT & 88.4 & 92.2 & 87.2 & 86.7 & 82.5 & 89.2 & 81.9 \\
PIXEL-base & 83.5 & 28.7 & 48.4 & 50.2 & 17.6 & 9.1 & 7.1 \\
PIXEL-bigrams & 64.5 & 15.9 & 25.5 & 23.1 & 11.8 & 7.3 & 11.1 \\
PIXEL-M4 & 79.0 & 30.4 & 52.7 & 32.5 & 40.3 & 27.8 & 17.3 \\
\bottomrule\end{tabular}
\caption{SIB-200 clean zero-shot macro F1 (\%). English-only fine-tuning uses all 701 training examples.}
\label{tab:sib-supplement}\end{table}

\begin{table}[t]\centering\scriptsize
\begin{tabular}{@{}lrrrrr@{}}\toprule
Model & EN & DE & FR & RU & ZH \\
\midrule
BERT & 73.9 & 24.3 & 31.5 & 23.4 & 23.4 \\
mBERT & 73.9 & 49.5 & 61.3 & 56.8 & 57.7 \\
XLM-R & 81.1 & 69.4 & 69.4 & 71.2 & 67.6 \\
PIXEL-base & 24.3 & 25.2 & 27.0 & 22.5 & 24.3 \\
PIXEL-M4 & 60.4 & 18.9 & 26.1 & 28.8 & 29.7 \\
\bottomrule\end{tabular}
\caption{Taxi1500 clean zero-shot accuracy (\%). The English source task has 860 training examples; several PIXEL checkpoints have weak English accuracy.}
\label{tab:taxi-supplement}\end{table}

\FloatBarrier
\end{document}